\documentclass[sn-aps]{sn-jnl}

\usepackage{enumitem}

\usepackage{array}

\usepackage{graphicx}%
\usepackage{multirow}%
\usepackage{amsmath,amssymb,amsfonts}%
\usepackage{amsthm}%
\usepackage{mathrsfs}%
\usepackage[title]{appendix}%
\usepackage{xcolor}%
\usepackage{textcomp}%
\usepackage{manyfoot}%
\usepackage{booktabs}%
\usepackage{algorithm}%
\usepackage{algorithmicx}%
\usepackage{algpseudocode}%
\usepackage{listings}%

\theoremstyle{thmstyleone}%
\theoremstyle{thmstyletwo}%

\theoremstyle{thmstylethree}%

\begin{document}

\title{Machine learning analysis of anomalous diffusion}


\author[1]{\fnm{Wenjie} \sur{Cai}}

\author[2]{\fnm{Yi} \sur{Hu}}

\author[1]{\fnm{Xiang} \sur{Qu}}

\author[1]{\fnm{Hui} \sur{Zhao}}

\author[1]{\fnm{Gongyi} \sur{Wang}}

\author[2]{\fnm{Jing} \sur{Li}}

\author*[1]{\fnm{Zihan} \sur{Huang}}\email{huangzih@hnu.edu.cn}

\affil[1]{\orgdiv{School of Physics and Electronics}, \orgname{Hunan University}, \orgaddress{ \city{Changsha}, \postcode{410082}, \country{China}}}

\affil[2]{ \orgname{Hubei Medical Devices Quality Supervision and Test Institute}, \orgaddress{ \city{Wuhan}, \postcode{430075}, \country{China}}}


\abstract{The rapid advancements in machine learning have made its application to anomalous diffusion analysis both essential and inevitable. This review systematically introduces the integration of machine learning techniques for enhanced analysis of anomalous diffusion, focusing on two pivotal aspects: single trajectory characterization via machine learning and representation learning of anomalous diffusion. We extensively compare various machine learning methods, including both classical machine learning and deep learning, used for the inference of diffusion parameters and trajectory segmentation. Additionally, platforms such as the Anomalous Diffusion Challenge that serve as benchmarks for evaluating these methods are highlighted. On the other hand, we outline three primary strategies for representing anomalous diffusion: the combination of predefined features, the feature vector from the penultimate layer of neural network, and the latent representation from the autoencoder, analyzing their applicability across various scenarios. This investigation paves the way for future research, offering valuable perspectives that can further enrich the study of anomalous diffusion and advance the application of artificial intelligence in statistical physics and biophysics.}

\keywords{ anomalous diffusion, machine learning, single trajectory characterization, representation learning}



\maketitle

\section{Introduction}\label{sec1}

Anomalous diffusion \cite{Metzler2014,Klafter2015,Manzo2023preface}, which refers to any type of diffusion that deviates from classical Brownian motion, is a fundamental phenomenon across a wide range of scientific disciplines, such as physics \cite{Mason1995,Aarao2016,Volpe2017,Barbosa2024,Iida2024,Hegde2022,Vitali2022,Shi2023,Zheng2022,Xu2021,Dai2022}, chemistry \cite{Muller1992,Ding2014,Barkai2012}, biology \cite{Wu2000,Illukkumbura2020,Gonzalez2008,Wang2013,Chen2016bio,Hofling2013,Zhang2021}, and finance \cite{Plerou2000,Masoliver2003,Jiang2019,Meyer2023}. As shown in figure 1(a), numerous experiments and observations have demonstrated that the occurrence of anomalous diffusion is ubiquitous across various scales and systems \cite{Munoz2021}. For instance, nanoparticles navigating through polymer networks with semiflexible strands perform enhanced heterogeneous diffusion [figure 1(b)] \cite{Xu2021, Dai2022}, while intracellular transport dynamics often exhibits distinct characteristic of anomalous diffusion [figure 1(c)] \cite{Zhang2021}. Even in ecological contexts, the seasonal migrations of species like springboks demonstrate non-standard movement patterns on a large scale [figure 1(d)] \cite{Meyer2023}. To comprehensively understand the mechanisms underlying these phenomena, the characterization and analysis of anomalous diffusion are crucial and increasingly demanded, providing key insights in the research of complex systems \cite{Sposini2022,Vilk2022sys,Wang2022,Timashev2010,Chen2013}.

The analysis of anomalous diffusion primarily emphasizes its distinctions from Brownian diffusion, with several characteristics typically identified. One notable characteristic is the nonlinearity of mean squared displacement (MSD):
\begin{equation}\label{MSD}
{\rm MSD}(t)\sim t^\alpha,
\end{equation}
where the diffusion exponent $\alpha$ categorizes the diffusion as either superdiffusive ($\alpha>1$) or subdiffusive ($\alpha<1$). Another distinct aspect involves the stochastic process in anomalous diffusion, which contrast significantly with the Wiener process associated with Brownian motion. For this reason, anomalous diffusion trajectories often exhibit long-range correlations and memory effects, or demonstrates the ``anomalous yet Brownian" diffusion characteristic \cite{Bowang2009,Mykyta2014,Haroldo2023,Haroldo2014,Yasmine2011,BoWang2012}. As a response, numerous theoretical models used to describe the evolution of anomalous diffusion have been proposed, such as the continuous-time random walk (CTRW) \cite{Scher,Dechant2019} and fractional Brownian motion (FBM) \cite{Mandelbrot}. CTRW stands out for its integration of random steps with variable waiting times, effectively modeling memory and non-Markovian effects. FBM extends traditional Brownian motion by incorporating a Hurst exponent, allowing it to describe random processes exhibiting persistent or anti-persistent characteristics. A third key distinction between anomalous diffusion and Brownian motion arises in complex environments, where a random walker may exhibit varying diffusion states over time due to intrinsic or extrinsic changes, leading to heterogeneous diffusion dynamics \cite{Chen2015,Persson2013,Monnier2015,Xu2021,Dai2022,Johnson1992,Cicerone1995,Yamamoto1998,Chepizhko2013,Yamamoto1998Heterogeneous,Lanoiselee2018,Hurtado2007,Cicerone1997,Cherstvy2013,weigel2011ergodic,manzo2015weak}. For instance, the Hfq protein in {\it Escherichia coli} cells can continuously transition among three diffusion states with different diffusion coefficients and mean dwell times \cite{Persson2013}. Similarly, in the microtubule network of DU145 cells \cite{Chen2015}, the transports of endosomes can vary among L\'{e}vy walk (LW) \cite{Klafter2} and Brownian motion. These complexities emphasize the need for advanced methods to decode the diffusion data and unravel the diverse nature of anomalous diffusion.

\begin{figure}[t]
\centering
\includegraphics[width=0.95\textwidth]{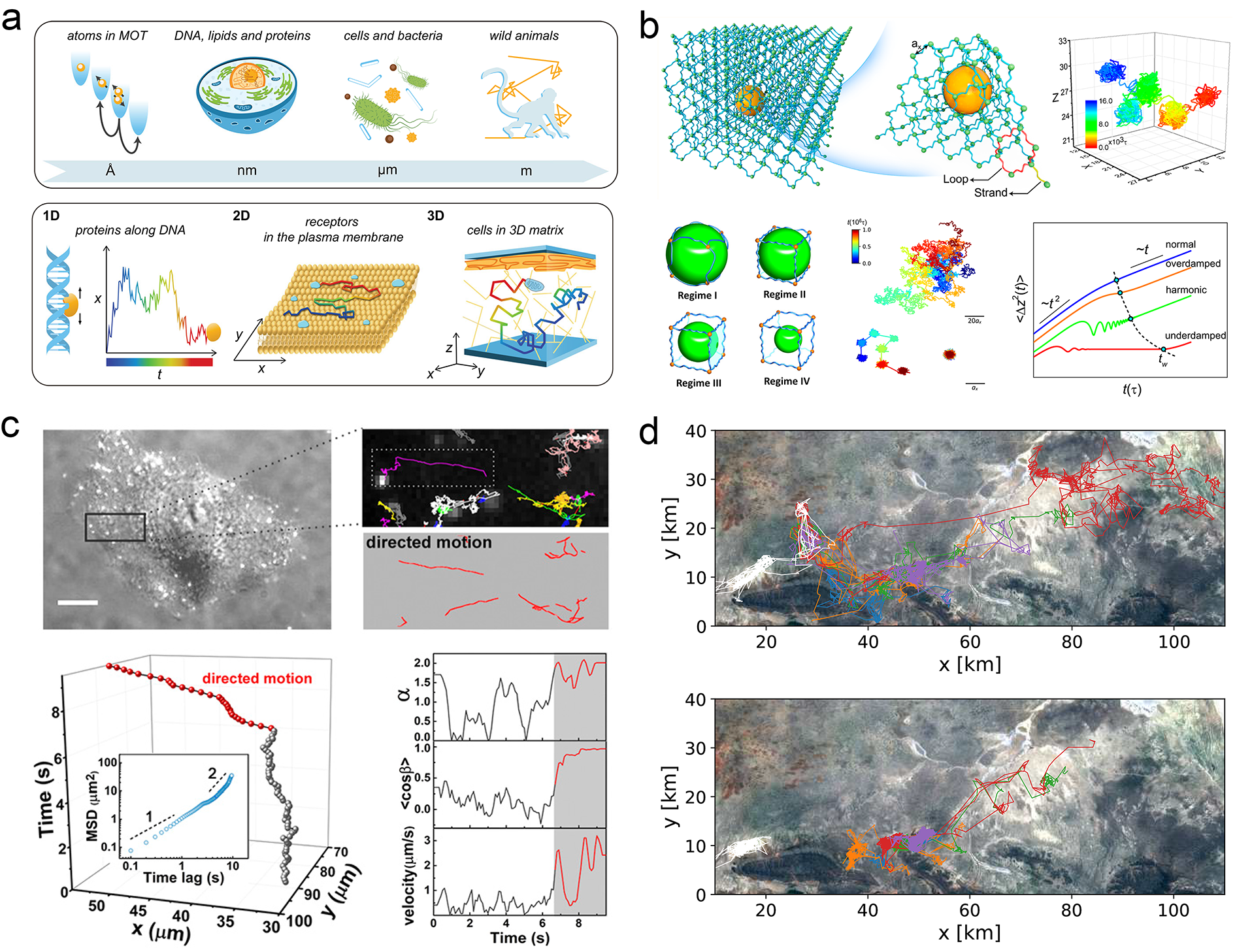}
\caption{(a) Illustrative images demonstrate anomalous diffusion across various length scales (top panel), and different dimensions (bottom panel), from \cite{Munoz2021}; (b) Enhanced heterogeneous diffusion can be identified for nanoparticles in a semiflexible polymer network, from \cite{Xu2021, Dai2022}; (c) Intracellular transport dynamics in early apoptotic cells, where trajectories of endocytic vesicles display typical anomalous diffusion properties, from \cite{Zhang2021}; (d) GPS-tracking movement data of springboks in Namibia during the wet season (top) and the dry season (bottom), both showing non-standard movement patterns on a large scale, from \cite{Meyer2023}.}
\label{fig:fig1}
\end{figure}

To address the challenges of analyzing anomalous diffusion, statistical methods \cite{Metzler2014,Metzler2019,Vilk2022,Sposini2022,Krapf2019,Burov2011,Sabzikar2022,Liu2017,Burnecki2015,Magdziarz2009,Meyer2022,Maraj2020,Magdziarz2020,Wang2020,Bhowmik2018,Katz1985, Tejedor2010, Ernst2014} are extensively used before the advent of machine learning techniques. These conventional approaches primarily relies on the construction of parametric features that effectively characterize the statistical properties of diffusion trajectories. For example, one key technique is the calculation of MSD, which can be used to infer the diffusion coefficients and diffusion exponents. Another is the self-part of the van Hove function \cite{Bhowmik2018} that quantifies the displacement distribution of a random walker. The non-Gaussianity of this function allows for a deeper understanding of the deviations from classical Brownian motion in anomalous diffusion behaviors. The strong interpretability of these statistical methods has led to their widespread use in the diffusion research, as they provide clear insights into the underlying mechanisms of diffusion processes.

However, inherent limitations of traditional statistical methods hinder their application, especially in modern research contexts. For example, these methods typically require long trajectories or large datasets to yield statistically reliable results. This requirement makes them less effective when dealing with short tracks or small datasets, which are often encountered in experimental settings. Another issue is their frequently inadequate accuracy, particularly in distinguishing between closely related diffusion behaviors under complex conditions. This is partly due to the reliance on assumptions that may not hold true across different systems or scales, such as the presumption of ergodicity or the application of homogeneous diffusion models to inherently heterogeneous environments. Moreover, these statistical methods are notably vulnerable to noise. This susceptibility can lead to potential misinterpretations of diffusion behaviors in trajectories with low signal-to-noise ratios, especially when experimental conditions are not ideal \cite{Seckler2023}. These limitations underscore the urgent demand for developing new analytical methods that can more effectively handle the diversity and complexity of anomalous diffusion data.

On the other hand, as an essential branch of artificial intelligence, machine learning techniques have undergone a rapid development in recent years \cite{He2016,Szegedy2015,Kiranyaz2021,Hochreiter1997,Zhai2018,Chen2016,Oord2016,Vaswani2017, Li2021cnn,Yu2019,Scarselli2008,Tavenard2020,Gruver2024,Worden2007,Ballard2021}. On the basis of big data, ML models excel in directly identifying patterns, making predictions, and informing decisions. In particular, ML is tremendously powerful in processing time-series data \cite{Tavenard2020,Gruver2024,Worden2007,Ballard2021}, which is critical for fields ranging from financial forecasting \cite{Tavenard2020} to health monitoring \cite{Worden2007} and environmental sensing \cite{Ballard2021}. The significance of ML in time-series analysis lies in its ability to model and predict complex sequential information without explicit programmatic definitions. ML algorithms can learn from historical data, allowing them to capture temporal dynamics and dependencies that are often missed by traditional statistical methods. This capability enables them to perform tasks such as anomaly detection, trend forecasting, and event prediction with high accuracy. Given the inherent nature of anomalous diffusion trajectories, which can be regarded as complex time-series data, ML is well-positioned to provide substantial analytical insights into these phenomena. It is foreseeable that ML algorithms will adeptly decode the subtle nuances in diffusion behavior by learning from the temporal sequence of data points, capturing long-range dependencies and patterns that may not be immediately obvious. Consequently, by leveraging ML's robust capabilities in time-series analysis, researchers can expect to obtain advanced tools for the analysis of diffusion processes.

Building upon these expectations, this review systematically summarizes the integration of ML techniques in the analysis of anomalous diffusion in recent years. The paper is organized as follows: We start with an examination of single trajectory characterization via ML in section 2, which focuses on the inference of diffusion parameters and segmentation of heterogeneous diffusion dynamics. ML competitions regarding anomalous diffusion analysis are also introduced in section 2. These competitions have catalyzed the development of a variety of high-performance ML algorithms specifically tailored for analyzing anomalous diffusion. Section 3 provides a comprehensive overview of research related to the representation learning of anomalous diffusion, discussing how to effectively represent the dynamics of anomalous diffusion. Finally, we summarize our findings and suggest promising research avenues in this domain in section 4.

\section{Single Trajectory Characterization via Machine Learning}\label{sec1}

The characterization of single trajectory properties is a fundamental step in providing detailed insights into the underlying physics of anomalous diffusion. Currently, the primary focus of single trajectory characterization centers around two critical tasks: the inference of diffusion parameters and the segmentation of heterogeneous diffusion dynamics, also known as trajectory segmentation. In this section, we will detail the applications of machine learning methods to these two tasks, along with the Anomalous Diffusion Challenge \cite{andi_website,Munoz2021}, which is a competition designed to benchmark the capabilities of various machine learning models against these tasks.

\subsection{Inference of diffusion parameters}\label{subsec2}

Accurately obtaining key diffusion parameters of trajectories is paramount in single trajectory characterization. Among the myriad diffusion parameters, the most representative parameters include the diffusion coefficient ($D$), the diffusion exponent ($\alpha$), and the theoretical diffusion model. The coefficient $D$ is a parameter that quantifies the diffusion speed, which can reflect properties such as the viscosity and temperature of the medium. The exponent $\alpha$ appears in the relationship between the MSD and time, revealing whether the diffusion is normal, subdiffusive, or superdiffusive. The diffusion model, such as CTRW and FBM, refers to the mathematical and physical framework used to theoretically describe the diffusion process. Based on the combined analysis of these parameters, researchers can understand the motion behaviors and dynamic mechanisms underlying the trajectory data.

Therefore, before the utilization of machine learning, numerous methods have been developed to infer these diffusion parameters. A classic example is the use of time-averaged mean squared displacement (TA-MSD) to determine the diffusion exponent $\alpha$ based on Eq. (\ref{MSD}). Another instance is the analysis of power spectral density (PSD), which provides the frequency-dependent information of the trajectories \cite{Metzler2019, Vilk2022, Sposini2022, Krapf2019}. In addition, velocity autocorrelation function (VACF) \cite{Burov2011}, fractional integral moving average (FIMA) process \cite{Sabzikar2022, Liu2017, Burnecki2015} and $p$-variation test \cite{Magdziarz2009} are also extensively employed when calculating the diffusion parameters. {Furthermore, Bayesian inference \cite{Thapa2018, Park2021, Thapa2022, Chen2022, Krog2017, Krog} has emerged as another powerful method for parameter estimation in the study of anomalous diffusion. It provides a probabilistic framework where prior knowledge about the system can be updated with observed data. This allows for continuous refinement of parameter estimates and offers a measure of uncertainty.} These traditional methods offer straightforward and highly interpretable analytical approaches that help us uncover the physical information embedded in diffusion trajectories.

However, intrinsic limitations are inevitable for statistics-based techniques. These traditional methods are generally sensitive to outliers and anomalies, and typically have a narrow scope when performing parameter inference. Each method is usually capable of inferring only one specific diffusion parameter, or distinguishing only two or few diffusion models. More importantly, with the rapid development of single particle tracking (SPT) technology \cite{Manzo2015spt, Shen2017, Qian1991, Saxton2008, Torreno-Pina2016, Elf2019, Cherstvy2019, Horton2010, Jeon2011, Leijnse2012, Codling2008, Gurtovenko2019,Smal2009, Roberts2020, Ye2023, Erimban2023, Javanainen2017, Winkler2018, Simon2024}, a growing array of diverse diffusion datasets has become available. Traditional statistics-based methods often struggle to effectively manage this complexity. These limitations underscore the urgent need to develop methods with robust generalization capabilities and the ability to handle complex datasets effectively.

To address this challenge, machine learning techniques have been introduced to the inference of diffusion parameters due to its powerful ability to handle complex data in recent years. In particular, inferring diffusion parameters essentially constitutes a regression or classification problem, fitting well within the standard supervised learning paradigm. For clarity in our discussion, we have organized these ML approaches into two primary categories in this section: classical machine learning and deep learning. Within classical machine learning, we highlight significant contributions using techniques like feature-based models \cite{Munoz-Gil2,Kowalek2022,Manzo2021,Loch2020}. Regarding deep learning, we focus on innovative applications that employ deep neural networks, including convolutional neural networks \cite{Gajowczyk2021,Granik2019,AL-hada2022,Conejero2023,Li2021,DeepSPT2020,NOA2020,Firbas2023,Feng2024}, recurrent neural networks \cite{Bo2019,Argun2021,Chen2022,Garibo-i-Orts2021,Kabbech2024}, graph neural networks \cite{Verdier2021, Verdier2022, Pineda2023}, to analyze diffusion trajectories and infer the parameters.

\subsubsection{Classical machine learning}\label{subsubsec2}

\vspace{\baselineskip}
Classical machine learning encompasses a range of algorithms and techniques that predate the extensive use of deep neural networks. These methods are grounded in established statistical learning theories and are characterized by their simplicity, interpretability, and lower computational requirements. For the analysis of anomalous diffusion, classical machine learning primarily relies on feature-based models to determine diffusion parameters.

As highlighted in figure 2(a), feature-based models {operate on feature vectors transformed from diffusion trajectories} and utilize classic machine learning algorithms such as decision trees and multilayer perceptrons for parameter inference. {The elements of these trajectory feature vectors are derived from data preprocessing, feature transformation, and feature extraction techniques.} By harnessing handcrafted features that capture the essential characteristics of the diffusion process, such as fractal dimension, maximal excursion, and mean gaussianity \cite{Katz1985, Tejedor2010, Ernst2014}, these models provide a structured way to analyze and interpret diffusion data. Typical related works include the following:

\begin{figure}
\centering
\includegraphics[width=0.95\textwidth]{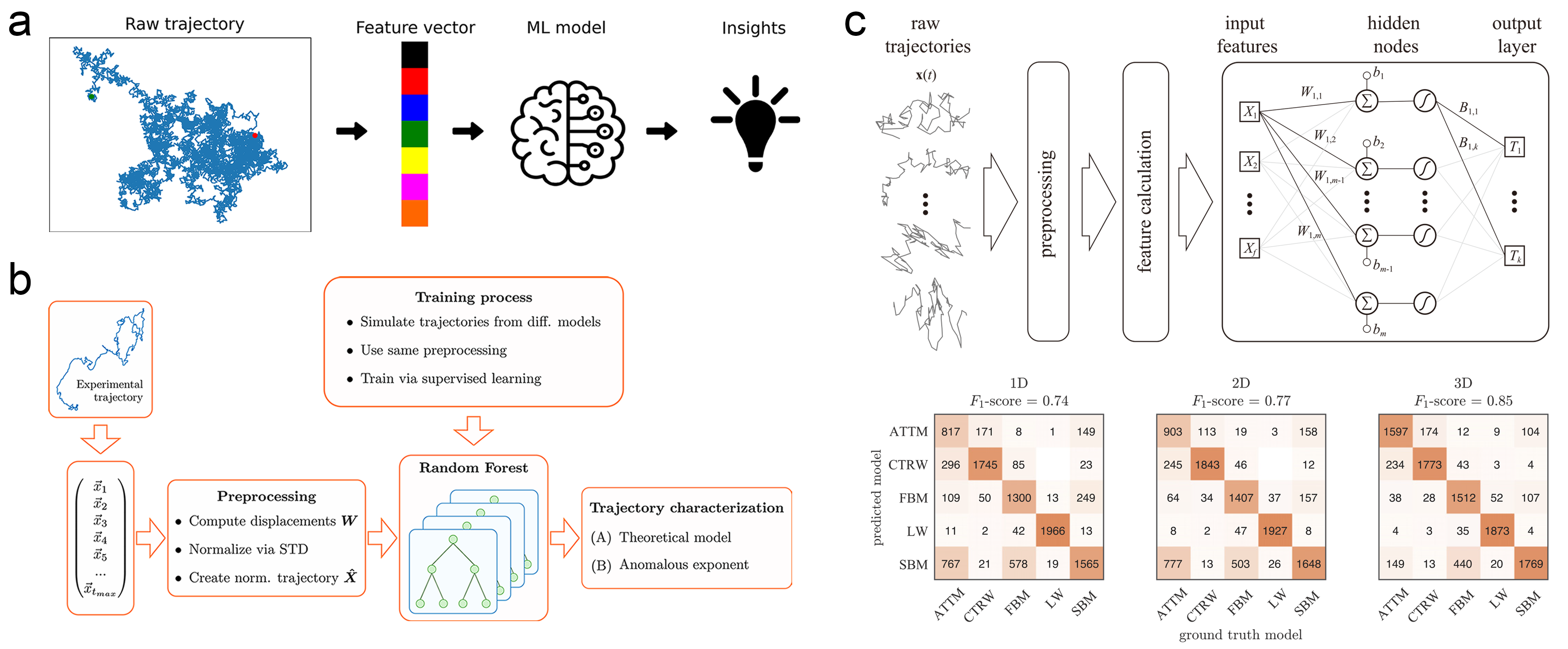}
\caption{(a) Schematic flowchart of the feature-based methods: a set of features is extracted from raw trajectories and used as the input for feature-based machine learning models, from \cite{Seckler2023}; (b) A workflow of random forest algorithm for the single trajectory characterization tasks, from \cite{Munoz-Gil2}; (c) {Schematic of the architecture of ELM (top), and corresponding performance in classifying diffusion models (bottom), from \cite{Manzo2021}.}}
\label{fig:fig2}
\end{figure}

\begin{itemize}
\item A study by G. Mu\~{n}oz-Gil et al. \cite{Munoz-Gil2} employs the random forest (RF) algorithm for the single trajectory characterization tasks, which is a classical machine learning method by aggregating results of multiple decision trees through voting or averaging As shown in figure 2(b), upon the supervision of labeled simulated trajectories from various diffusion models, the RF model can be trained to infer diffusion exponent and theoretical model after {preprocessing of these trajectories}. The trained model is then applied to experimental data, including the movement of mRNA molecules in bacterial cells and membrane receptor diffusion in living cells. The results show that the RF model achieves performance superior to traditional methods, particularly in the analysis of nonergodic diffusion processes.

\item The work by P. Kowalek et al. \cite{Kowalek2022} utilizes the gradient boosting algorithm to analyze the diffusion trajectories. This algorithm is an ensemble machine learning technique that progressively refines models by addressing errors from previous iterations. Each subsequent model is specifically trained to minimize the residuals left by its predecessors, culminating in a prediction that represents a weighted amalgamation of all successive models. In particular, the authors choose XGBoost \cite{Chen2016} in this work, which is a decision-tree-based version of gradient boosting algorithm. Initially, the study involves extraction and selection of diverse features from raw diffusion trajectories, utilizing the XGBoost to categorize different diffusion behaviors. This process is iteratively refined by integrating new features, significantly enhancing the classifier's accuracy. The robustness of this method is further validated by its application to complex diffusion patterns, illustrating its potential in providing insightful analysis of anomalous diffusion dynamics from various perspectives.

\item Another application of classical machine learning is the use of extreme learning machine (ELM) in C. Manzo's work \cite{Manzo2021}. This method is tailored for single-layer feedforward networks [figure 2(c)], simplifying the learning process by randomly assigning weights between the input and hidden layers. In this study, ELM is integrated with feature engineering to analyze anomalous diffusion trajectories. The input to the ELM model includes features such as time-normalized mean absolute displacement, mean square displacement, and cumulative square displacement. These features are efficiently processed by ELM to infer critical diffusion parameters like diffusion exponent and diffusion model. The model performance underscores the effectiveness of ELM in handling complex, feature-rich datasets, demonstrating its potential to deliver rapid and accurate analysis of diffusion behaviors.
\end{itemize}

In practice, the effectiveness of these feature-based models heavily depends on the {features transformed from trajectory data}, such as TA-MSD, displacement moments, or other engineered features that characterize the diffusion dynamics, which will be discussed in detail in section 3.1. As these features typically possess strong statistical or physical significance, they provide feature-based methods with enhanced interpretability compared to deep learning. However, selecting the optimal combination of features is challenging, which requires deep domain expertise and extensive experimental analysis to identify the most significant characteristics of diffusion properties. In this regard, H. Loch-Olszewska et al. underscores the impact of feature choice on machine learning classification of anomalous diffusion \cite{Loch2020}. The study shows that while combining multiple feature sets can enhance the performance of classifier, no single feature set can universally excel across all diffusion models. This variability highlights the profound impact of feature selection, and also suggests that feature-based machine learning methods may have limitations in their universality for the analysis of anomalous diffusion.

\subsubsection{Deep learning}\label{subsubsec2}

While classical machine learning techniques have demonstrated substantial utility in analyzing anomalous diffusion, the advent of deep learning offers a transformative approach due to its inherent capability to manage big data with complex patterns. Unlike feature-based methods that require manual feature extraction, deep learning strategies prefer integrating raw trajectory data directly into deep neural network architectures. This shift enables a more granular understanding of diffusion processes by leveraging minimal preprocessing, maintaining focus on the intrinsic temporal sequences of trajectory data. Building on this advanced methodology, deep learning models such as convolutional neural network (CNN) \cite{Li2021cnn, Gajowczyk2021, Granik2019,AL-hada2022,Conejero2023,Li2021,DeepSPT2020,NOA2020,Firbas2023,Feng2024}, recurrent neural network (RNN) \cite{Yu2019, Bo2019, Argun2021, Garibo-i-Orts2021, Chen2022,Kabbech2024}, and graph neural network (GNN) \cite{Scarselli2008, Verdier2021, Verdier2022, Pineda2023} are at the forefront of analyzing anomalous diffusion.

CNN stand as a fundamental neural network architecture in deep learning, particularly renowned for its ability to process spatial and temporal data like images and time series. It excels in autonomously extracting hierarchical features from raw data, which is crucial for comprehensive analysis across various scales. Pioneering this approach in anomalous diffusion, the study by N. Granik et al. \cite{Granik2019} shows the application of CNNs in the classification and parameter estimation of single particle trajectories. As shown in the top of figure 3(a), the authors transform the 2D trajectory data into a numerical matrix as the input. Subsequently, by employing a CNN where the layers incrementally increase dilation factors, this model adeptly predicts the motion types and captures correlations of trajectory data across extensive temporal spans. Another study conducted by E. A AL-hada et al. \cite{AL-hada2022} explores the adaptability of CNNs in discerning complex stochastic processes, where the input is the time-evolution image of a trajectory [bottom of figure 3(a)]. This research utilizes an array of pre-trained CNN models, including ResNet-18, ResNet-50 \cite{He2016}, and GoogLeNet \cite{Szegedy2015}, to delve into various stochastic categories like CTRW, FBM, LW, and multi-state processes. The results demonstrate the superior capability of CNNs in classifying these stochastic behaviors with high accuracy. In addition, the use of 1D CNN \cite{Kiranyaz2021, Li2021,Conejero2023,Firbas2023,Gajowczyk2021} enables the model without the need for additional trajectory preprocessing, enhancing the CNN's ability to handle raw trajectory data directly. These works collectively illustrate the effectiveness of CNNs in inferring parameters of anomalous diffusion.

\begin{figure}
\centering
\includegraphics[width=0.95\textwidth]{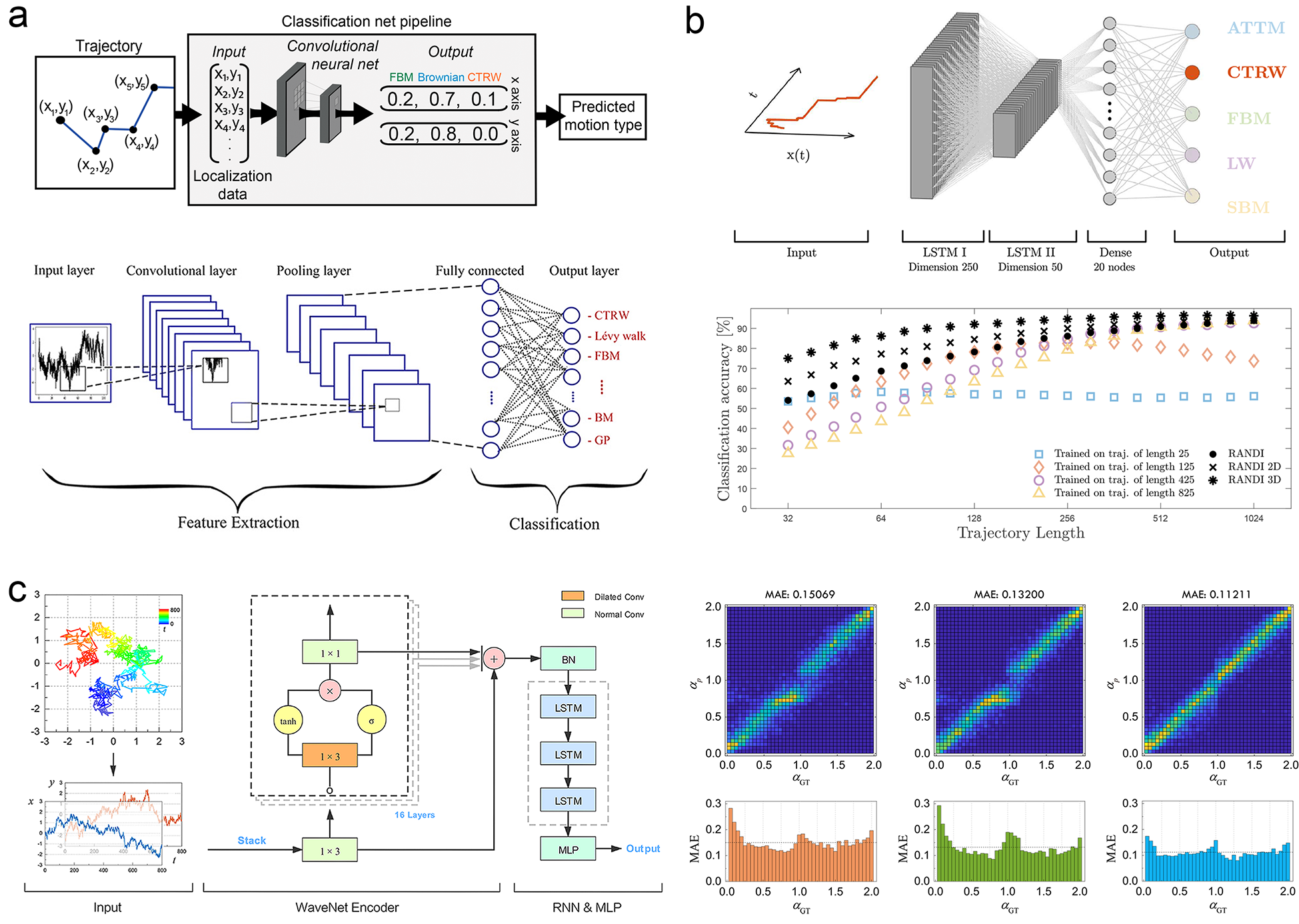}
\caption{(a) Schematic diagrams of single trajectory characterizations using CNNs, from \cite{Granik2019, AL-hada2022}; (b) The workflow (top) and model performance (bottom) of RANDI model, which has a dual-layer LSTM structure, from \cite{Argun2021}; (c) Workflow (left) and performance analysis (right) of WADNet, which is combined by a WaveNet encoder and a 3-layer LSTM network, from \cite{Li2021}.}
\label{fig:fig3}
\end{figure}

Compared to the diverse preprocessing requirements of CNNs for trajectory data, RNNs are inherently suited for time-series analysis, allowing them to process raw trajectory data with minimal preprocessing. On the other hand, the memory capabilities of RNNs, especially long short-term memory (LSTM) networks, enable them to effectively analyze the long-range correlations within trajectories. A notable application of RNNs in this domain is the work by S. Bo et al. \cite{Bo2019}, where the LSTM network is leveraged to infer the diffusion exponent from experimental trajectory data of sub-diffusive colloids and super-diffusive microswimmers. This work underscores the ability of RNNs to process the intricate dynamics of such systems, especially in environments with speckle light fields where traditional methods falter. Further expanding on this, the RANDI model, developed by A. Argun et al. \cite{Argun2021}, incorporates a dual-layer LSTM structure [figure 3(b)], refining the model's ability for the inference of diffusion parameters. These studies \cite{ Bo2019, Argun2021, Garibo-i-Orts2021, Chen2022,Kabbech2024} illustrate RNNs' potential to provide deep insights into the underlying physics of anomalous diffusion.

Moreover, following the successful applications of CNNs and RNNs in anomalous diffusion, researchers have begun experimenting with hybrid architectures that combine different types of machine learning models to achieve better performance. For instance, as shown in figure 3(c), D. Li et al. \cite{Li2021} propose a deep learning architecture called WADNet, which has an excellent ability for the diffusion parameter inference. This model combines an improved WaveNet \cite{Oord2016} encoder (a specialized 1D CNN) with a 3-layer LSTM network to infer diffusion exponents and classify diffusion models. In particular, WADNet achieves outstanding performance in the $1^{\rm st}$ Anomalous Diffusion Challenge \cite{andi_website}, surpassing the first places in the competition leaderboard across all tasks and dimensions. In addition to this, there have been many combination attempts, such as combining ResNet with XGBoost \cite{DeepSPT2020}, CNN with bi-LSTM \cite{NOA2020}, and Bayesian inference with deep networks \cite{Seckler2022}. These strategies not only integrate the strengths of various networks but also compensate for the shortcomings of individual models, making them a more effective method for the analysis of anomalous diffusion.

\begin{figure}
\centering
\includegraphics[width=0.95\textwidth]{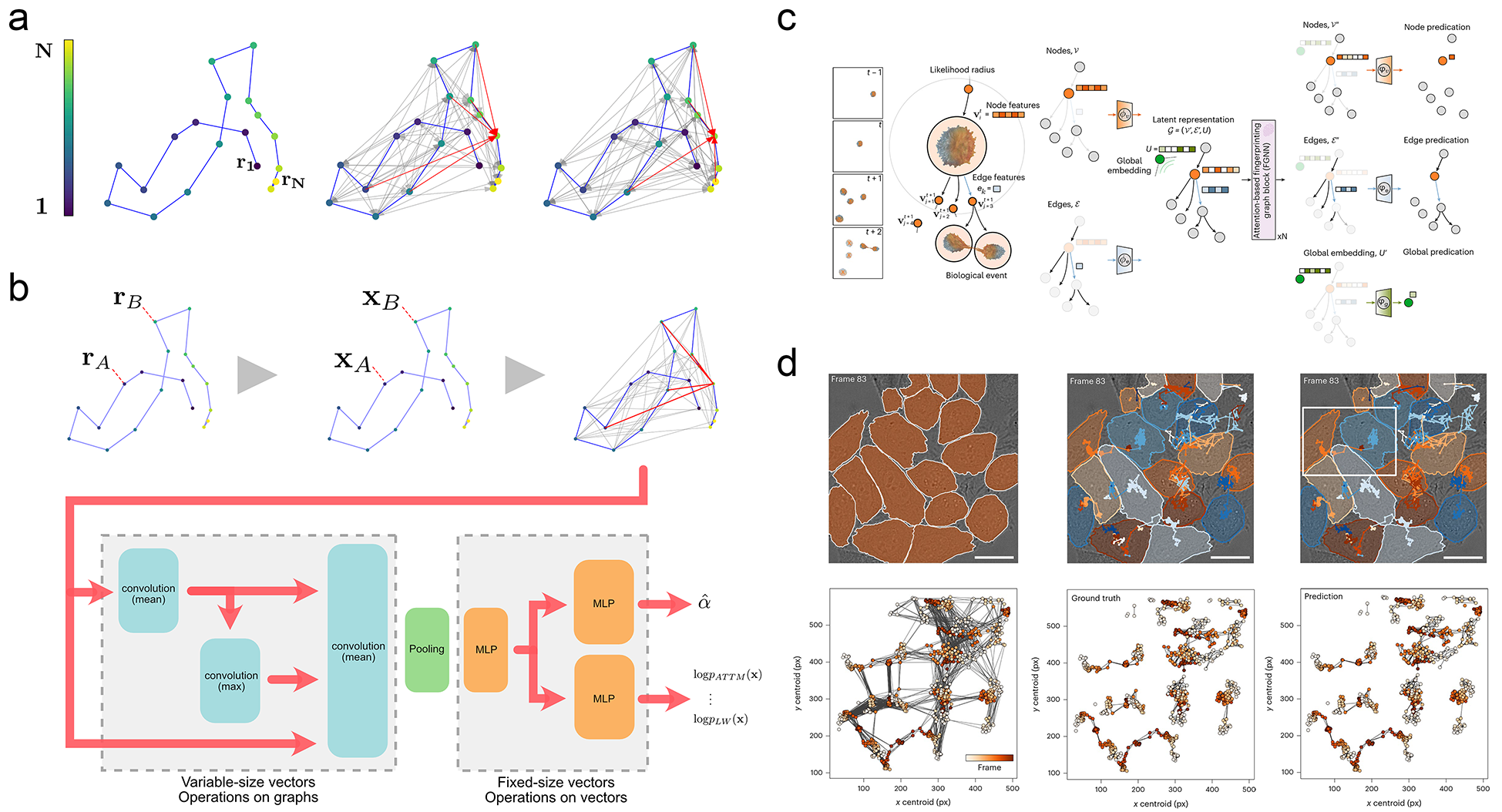}
\caption{(a) Representative example of graphs associated with a short diffusion trajectory, from \cite{Verdier2021}; (b) Workflow of the GNN for processing diffusion trajectories, from \cite{Verdier2021}; (c) Schematic diagram showing the spatiotemporal graph representation of trajectories using MAGIK, from \cite{Pineda2023}; (d) Comparative analysis of trajectory linking results using MAGIK on HeLa cell video data with ground truth trajectories, from \cite{Pineda2023}.}
\label{fig:fig4}
\end{figure}

Furthermore, GNN offers a distinct advantage for analyzing anomalous diffusion by leveraging their unique ability to process graph-structured data.  This capability is essential when representing diffusion trajectories as complex graphs, where nodes symbolize discrete trajectory points and edges encapsulate the temporal or spatial relationships between these points. GNNs excel in handling these non-Euclidean structures through a message-passing mechanism that integrates multi-scale information across various nodes. By employing iterative aggregation, GNNs capture both immediate and extended temporal dependencies, providing a deep understanding of the complex relationships inherent in diffusion processes. For instance, in the work by H. Verdier et al. \cite{Verdier2021} , the GNN model is structured to transform each trajectory into a graph, as illustrated in figures 4(a) and 4(b). The input graph is processed through several graph convolution operations that project it into a latent space [figure 4(b)]. This representation facilitates the detailed estimation of the anomalous exponent and classifies the type of random walk effectively. Another example is the development of MAGIK framework by J. Pineda et al. \cite{Pineda2023}. MAGIK utilizes a geometric deep learning approach that models motion through graph representations, employing attention-based components for dynamic estimation [figure 4(c)]. This framework showcases flexibility across diverse biological experiments [figure 4(d)], proving GNNs as a robust tool for spatiotemporal analysis without the need for explicit trajectory linking. These robust applications of GNNs \cite{Verdier2021, Verdier2022, Pineda2023} highlights  their potential to significantly enhance our understanding of complex stochastic processes.

\subsubsection{Further discussions}\label{subsubsec2}

While both classic machine learning and deep learning have demonstrated outstanding performance in the inference of diffusion parameters, there are critical considerations that need to be addressed for these methods to reach their full potential. First, the application of machine learning in this field primarily falls within the paradigm of supervised learning, where the task is either a classification or regression problem. Therefore, large and high-quality datasets with accurate labels are required when training these models. However, obtaining labels from experimental data can be challenging. To address this issue, the main current approach is the {\it simulation-based} method \cite{Munoz-Gil2,Kowalek2022,Manzo2021,Loch2020,Seckler2022,Granik2019,AL-hada2022,Conejero2023,Gajowczyk2021,Li2021,Firbas2023,DeepSPT2020,NOA2020,Bo2019,Argun2021,Chen2022,Garibo-i-Orts2021,Kabbech2024,Verdier2021,Pineda2023,Verdier2022, QuSoftware}. This method uses various known mathematical diffusion models, such as CTRW, FBM, and LW, to simulate trajectories. These trajectories can be labeled based on the diffusion models and corresponding simulation parameters. After training the machine learning models on simulated trajectories, these models can be subsequently applied to the analysis of real experimental data. {In particular, the first Anomalous Diffusion Challenge \cite{Munoz2021,andi_website} specifically tests several competing simulation-based methods on real experimental data \cite{golding2006physical,manzo2015weak,stadler2017non,kindermann2017nonergodic,Krapf2019}, like the trajectories of mRNA molecules inside live {\it E. coli} cells \cite{golding2006physical} and single atoms moving in a 1D periodic optical potential \cite{kindermann2017nonergodic}. The top-performing methods in this competition show consistent predictions and agree well with experimental values, highlighting that simulation-based approaches can produce robust results across different experimental settings. Details of this competition will be introduced in section 2.3.} Although the simulation-based method has shown excellent performance so far, it faces an inherent challenge due to the natural differences between simulated and experimental data. On one hand, experimental trajectories often exhibit higher levels of noise, stochasticity, and complexity compared to their simulated counterparts. On the other hand, many characteristics of experimental trajectories cannot be fully captured and simulated by known diffusion models. This clearly limits the generalizability of machine learning models trained solely on synthetic data. Thus, the improvement and optimization of the simulation-based method will be a critical issue for future research.

Second, each of classical machine learning and deep learning has its own strengths and limitations when applied to the inference of diffusion parameters. Understanding these differences is crucial for selecting the appropriate method based on the research objectives. In the following, we compare the characteristics of these two types of machine learning models from three perspectives: input form, interpretability, and nonlinear modeling capabilities.

\begin{enumerate}

\item The forms of input data significantly diverge between classical machine learning and deep learning techniques. Classical machine learning depends heavily on preprocessing and feature engineering. This task often necessitates considerable domain knowledge and can be labor-intensive. The effectiveness of these models is largely contingent upon the quality and appropriateness of the chosen features, making them particularly effective when the features are tailored to the anomalous diffusion. Conversely, deep learning methods excel in processing raw data, largely circumventing the need for manual feature engineering. These models can autonomously learn to identify relevant features during their training processes. This direct data consumption enables deep learning models to be more adaptable and scalable, advantageous in complex scenarios or when the critical features of the data are unknown.

\item The interpretability of models represents a crucial distinction between classical machine learning and deep learning approaches. Classical techniques like decision trees and random forests offer higher transparency due to their reliance on explicitly defined features. As stated before, these crafted features are selected with substantial domain knowledge. They often root in well-understood physical or statistical properties of the diffusion process, such as TA-MSD and VACF. {Therefore, these feature-based representations of anomalous diffusion often prove more robust to differences between experimental and simulation data, due to their more generalized encoding of the trajectory.} This level of clarity is invaluable, especially in scientific research where understanding the underlying mechanisms is as crucial as the predictive performance of the model. On the contrary, deep learning models are often criticized for their ``black box" nature, as their internal workings remain largely opaque. {The lack of interpretability makes it difficult to verify the predictions of deep learning models, sometimes leading to false confident predictions. This is also highlighted in Ref. \cite{Feng2024} when dealing with the out-of-distribution dynamics of anomalous diffusion.} Nevertheless, it is possible to gain insights into the role of each component within deep neural networks through methods such as ablation studies \cite{Meyes2019,Qu2024}. These techniques can help elucidate the underlying mechanisms by which neural networks infer diffusion parameters.

\item The capacity to model nonlinear relationships is another essential distinction between classical machine learning and deep learning. Classical approaches often rely on linear assumptions or straightforward nonlinear transformations, thereby emphasizing the importance of feature engineering. By contrast, deep learning is inherently adept at capturing complex nonlinear interactions due to its multilayered architecture. This profound capability to model nonlinearities is crucial when dealing with anomalous diffusion. This is because the nonlinear information within anomalous diffusion trajectories, such as long-range correlations and frequency space information, is often implicit and challenging to fully uncover through manual feature extraction. In particular, deep learning can autonomously discover these hidden complex nonlinear patterns. Although these patterns remain implicit within neural networks, this does not hinder deep learning from being a potent tool in characterizing anomalous diffusion.
\end{enumerate}

\subsection{Segmentation of heterogeneous diffusion dynamics}\label{sec2}

As highlighted in the introduction, a significant distinction between anomalous diffusion and Brownian motion emerges in complex environments, where a random walker may exhibit varying diffusion states over time due to intrinsic or extrinsic changes, leading to heterogeneous diffusion dynamics \cite{Chen2015,Persson2013,Monnier2015,Xu2021,Dai2022,Johnson1992,Cicerone1995,Yamamoto1998,Chepizhko2013,Yamamoto1998Heterogeneous,Lanoiselee2018,Hurtado2007,Cicerone1997,Cherstvy2013,weigel2011ergodic,manzo2015weak}. Compared to single-state trajectories with fixed properties, heterogeneous dynamics involve transitions among multiple diffusion states. This makes the accurate analysis of multi-state trajectory data much more complex and thus poses significant challenges in the single trajectory characterization. Segmentation of heterogeneous diffusion dynamics is therefore a crucial task in the analysis of anomalous diffusion. This task involves two primary subtasks:
\begin{enumerate}
\item Identifying transition points where changes in the diffusion state occur. We call these transition points as {\it changepoints} in this work.

\item Determining the diffusion parameters for each segmented sub-trajectory. For convenience, we use the term {\it segments} to refer to these segmented sub-trajectories.
\end{enumerate}
The combination of these two complex subtasks leads to a diverse range of techniques for segmenting trajectories of heterogeneous diffusion dynamics. In this article, we primarily introduce two mainstream types of segmentation approaches: the sliding window and point-wise methods.

\begin{figure}
\centering
\includegraphics[width=0.95\textwidth]{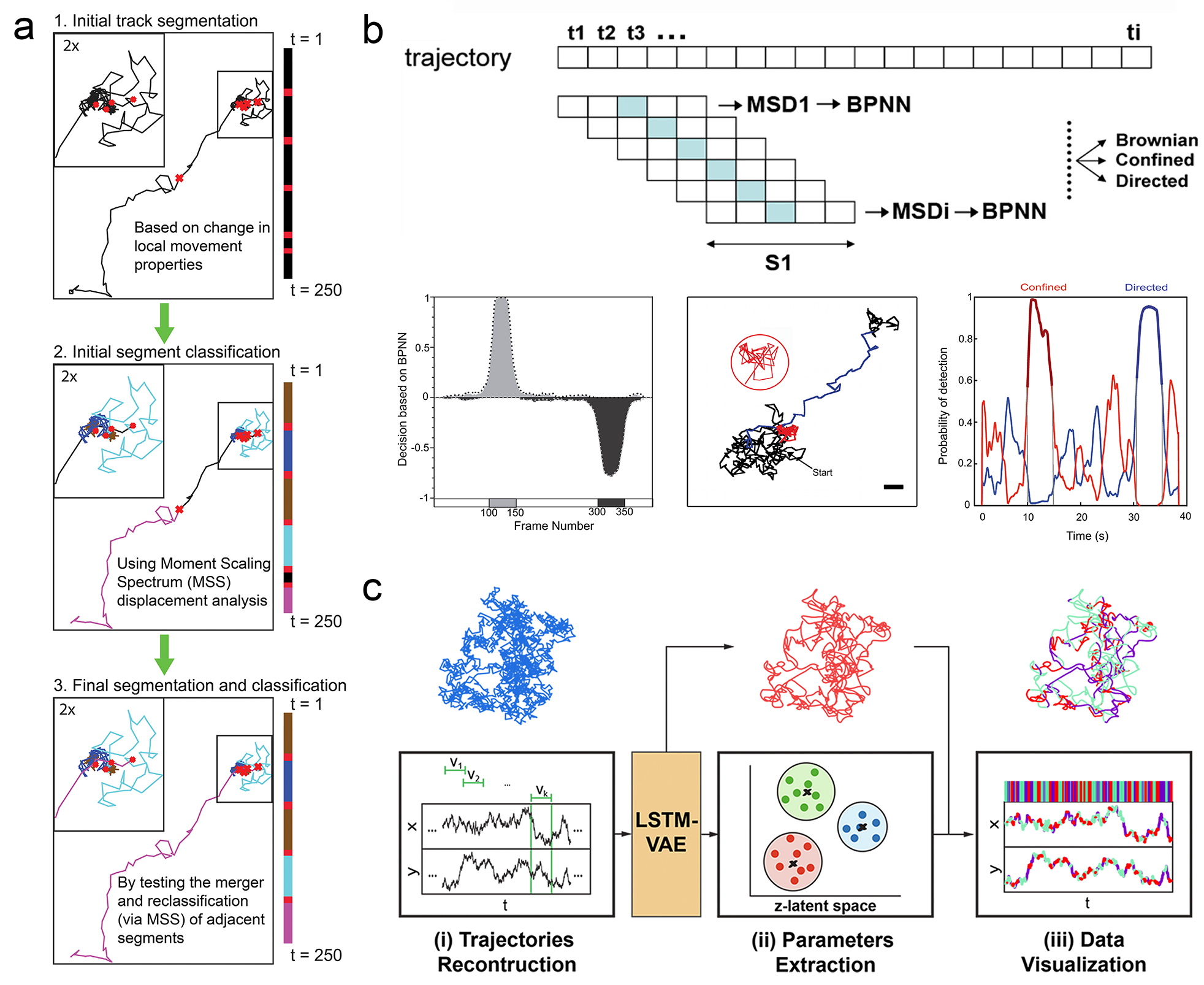}
\caption{(a) Illustration of three segmentation steps in the DC-MSS algorithm, from \cite{Vega2018}; (b) The combination of sliding window method and BPNN for trajectory segmentation (top) and corresponding performance analysis (bottom) , from \cite{Dosset2016}; (c) Workflow of Deep-SEES for trajectory segmentation where the sliding window and LSTM-VAE network are utilized, from \cite{Zhang2023}.}
\label{fig:fig5}
\end{figure}

\subsubsection{Sliding window method}\label{subsubsec2}

As the name suggests, the sliding window method involves segmenting a trajectory progressively using a sliding window algorithm \cite{Yu2014}. The window length is predefined, dividing the trajectory into several segments of equal length. After that, a classifier is employed to predict the diffusion model associated with each segment. This classifier can be either a traditional statistical method or a machine learning model. By detecting the abrupt changes in predictions for consecutive segments, the changepoints of diffusion states can be identified.

Before the application of machine learning to the analysis of anomalous diffusion, researchers have already employed the sliding window method to segment multi-state diffusion trajectories. A notable example is the divide-and-conquer moment scaling spectrum (DC-MSS) approach developed by A. R. Vega et al. \cite{Vega2018}. As depicted in figure 5(a), DC-MSS begins with an initial segmentation using an 11-frame sliding window to identify potential changepoints. This is followed by the classification of these segments through moment scaling spectrum analysis, which categorizes them into different motion types. The final step involves refining the segmentation by merging and reclassifying adjacent segments to ensure accurate changepoint identification. This method is tested on tracking data of the transmembrane protein CD44 on the surface of macrophages, revealing transitions between various motion types, including free diffusion, confined diffusion, directed diffusion, and immobilization.

While the sliding window method permits trajectory segmentation using traditional statistical approaches, these approaches often fall short when faced with more complex trajectory data. Consequently, integrating machine learning techniques with the sliding window method has enabled the creation of models that possess enhanced segmentation capabilities. An illustrative instance of this integration is the algorithm developed by P. Dosset et al. \cite{Dosset2016}, which employs a back-propagation neural network (BPNN) in conjunction with a 31-frame sliding window to classify diffusion models. As shown in figure 5(b), BPNN performs localized MSD analysis on segments delineated by the sliding window, enabling the detection of transitions among distinct diffusion models such as Brownian, confined, and directed motions. This algorithm has been tested on experimental trajectories of transmembrane proteins and viral particles, validating its capacity to accurately segment multi-state diffusion trajectories.

Another application of this technique is the Deep-SEES method introduced by Y. Zhang et al. \cite{Zhang2023}. Utilizing a sliding window approach, this method segments trajectories for subsequent analysis with a long short-term memory with variational autoencoder (LSTM-VAE) network, as illustrated in figure 5(c). This network excels at extracting latent features and classifying dynamic states, addressing challenges like noise and undersampling effectively. This segmentation capability is crucial for identifying rare events, such as when nanoparticles cross the membrane, which may be hidden within complex or noisy trajectories. Specifically, in active enzyme systems, Deep-SEES adeptly segments the heterogeneous dynamics of gold nanorods. Similarly, in liquid-liquid phase separation systems, it accurately captures the distinct rotational behaviors of nanoparticles. Additionally, other machine learning models such as support vector machine (SVM) \cite{Helmuth2007} and K-means clustering \cite{Weron2017, Sikora2017}, when integrated with the sliding window method, have also demonstrated effectiveness in trajectory segmentation. For a detailed overview of such integrations and their features, refer to the selected examples listed in Table 1. These integrations further broaden the application spectrum of sliding window methods, offering diverse analytical tools capable of addressing the segmentation of heterogeneous diffusion dynamics.

\subsubsection{Point-wise method}\label{subsubsec2}

While the sliding window method offers a systematic approach to segmenting trajectories, it inherently suffers from limitations, particularly regarding the predefinition of window length. {Considering that the performance of machine learning algorithms heavily depends on the trajectory length \cite{Bo2019,Li2021,Munoz2021,Verdier2022,Firbas2023}}, the choice of window length is often a trade-off. {Longer trajectories provide richer, more comprehensive data, which can be advantageous for capturing overall diffusion patterns. However, using larger windows may smooth out rapid transitions between states, potentially missing important short-term dynamics. Conversely, shorter trajectories tend to lack sufficient information, making it difficult for machine learning models to extract meaningful features. In these cases, smaller windows might increase noise, which can obscure the underlying diffusion process and lead to misclassifications.} These constraints become more pronounced when dealing with complex and heterogeneous diffusion dynamics, where transitions between states do not necessarily follow uniform lengths.

\begin{figure}
\centering
\includegraphics[width=0.95\textwidth]{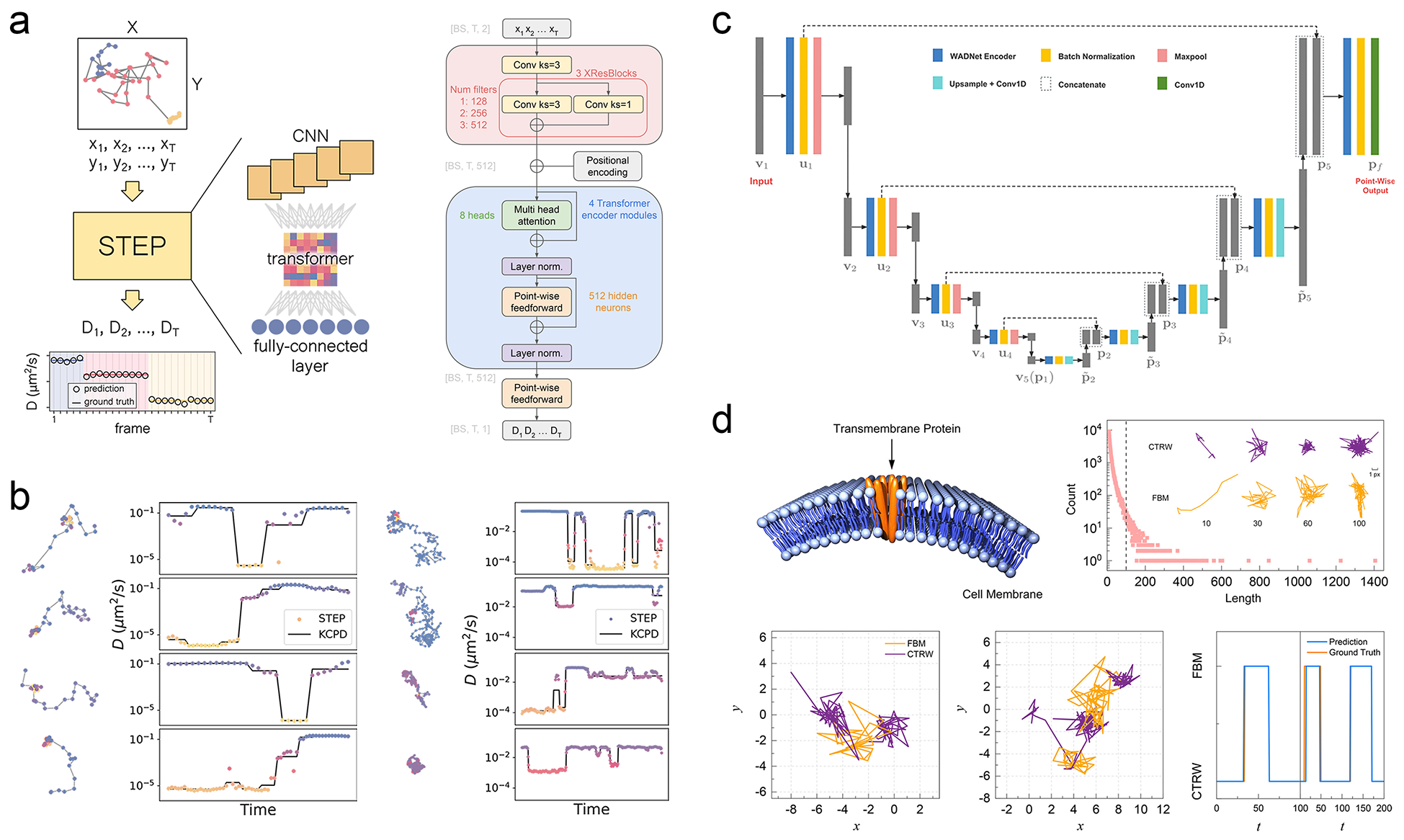}
\caption{(a) Schematic diagram of the STEP model, from \cite{Requena2023}; (b) Segmentation results of diffusion coefficients given by STEP for experimental trajectories of integrin $\alpha5\beta1$ (left) and the pathogen-recognition receptor DC-SIGN (right), from \cite{Requena2023}; (c) Workflow of the U-AnDi model, from \cite{Qu2024}; (d) Segmentation performance of U-AnDi on the diffusion of transmembrane proteins on cell membrane surfaces, from \cite{Qu2024}.}
\label{fig:fig6}
\end{figure}

\begin{sidewaystable}
\raggedleft
\caption{Overview of selected methods for the segmentation of anomalous diffusion {(partially derived from Ref. \cite{Munoz2023}).} }\label{tab:segmentation_methods}
\begin{tabular*}{\textheight}{@{\extracolsep\fill}p{9cm}cccc}
\toprule%
Method Name & Method Type & Learning Type & Sim.-Based & Seg. Target  \\
\midrule
  Divide-and-conquer moment scaling spectrum (DC-MSS) \cite{Vega2018} & Sliding window & Stat. & No & model \\

  Sliding-window time-averaged MSD \cite{Weron2017} & Sliding window & Stat. & No & $\alpha$\\

  Recurrence analysis with sliding window segmentation \cite{Sikora2017} & Sliding window & Stat. & No & model \\

  Probability density function and autocorrelation function analysis \cite{Matsuoka2009} & Sliding window & Stat. & No & $D$ \\

  Extreme learning machine (AnDi-ELM) \cite{Manzo2021} & Sliding window & ML & Yes & model, $\alpha$ \\

  Recurrent neural networks$^{a}$ \cite{Bo2019} & Sliding window & ML & Yes & $\alpha$ \\

  Recurrent neural networks$^{b}$ (RANDI) \cite{Argun2021} & Sliding window & ML & Yes & model, $\alpha$ \\

  Back-propagation neural network with MSD analysis \cite{Dosset2016} & Sliding window & ML & Yes & model\\

  Extracting features of subtrajectories with historical experience learning (Deep-SEES) \cite{Zhang2023} & Sliding window & ML & Yes & model \\

  Multi-feature random forest trajectory analysis \cite{Wagner2017} & Sliding window & ML & Yes & model \\

  Bayesian model selection to hidden Markov modeling \cite{Monnier2015} & Sliding window & ML+Stat. & No & model, $v$ \\

  Gamma mixture and hidden Markov model \cite{Matsuda2018} & Sliding window & ML+Stat. & No & $D$ \\

  Feature-engineered classical statistics combined with supervised deep learning (CONDOR) \cite{Gentili2021} & Sliding window & ML+Stat. & Yes & model, $\alpha$\\

  Deep learning followed by moment scaling spectrum analysis (DL-MSS) \cite{Arts2019} & Sliding window & ML+Stat. & Yes & model \\

  Geometric deep learning framework with attention mechanisms (MAGIK) \cite{Pineda2023} & Point-wise & ML & Yes & model, $\alpha$, $D$ \\

  Deep convolutional networks (U-AnDi) \cite{Qu2024} & Point-wise & ML & Yes & model, $\alpha$ \\

  Point-wise diffusion property prediction (STEP) \cite{Requena2023} & Point-wise & ML & Yes & $D$, $\alpha$ \\

  Recurrent neural networks$^{c}$ \cite{Martinez2023} & Point-wise & ML & Yes & model \\

  Nonparametric Bayesian inference (NOBIAS) \cite{Chen2022} & Point-wise & ML+Stat. & Yes & model \\

  {Bayesian deep learning} \cite{seckler2024change} & {Point-wise} & {ML+Stat.} & {Yes} & {model} \\
\botrule
\end{tabular*}
\footnotetext{Note: Meanings of new symbols and abbreviations in the table are listed as follows. $v$: velocity. Stat.: statistics. Sim.: simulation. Seg.: segmentation.}
\end{sidewaystable}

To mitigate issues related to fixed segment lengths, researchers have developed the point-wise method, which is an approach inspired by the semantic segmentation in computer vision \cite{Mo2022}. Semantic segmentation, also known as pixel-wise segmentation in image processing, classifies each pixel into specific categories. This technique enhances object recognition and delineates boundaries within images with high precision. Analogously, the point-wise method in trajectory analysis considers each data point along a trajectory as a potential changepoint. This allows for a more fine-grained and adaptable identification of state transitions, offering a distinct advantage in scenarios where diffusion states may change rapidly or irregularly.

One representative application of the point-wise technique is the STEP method developed by B. Requena et al. \cite{Requena2023}. As shown in figure 6(a), this method leverages deep learning models, including CNN and Transformer \cite{Vaswani2017} encoder, to segment the trajectory in a point-wise manner. This design allows the model to extract critical diffusion features such as diffusion coefficients and diffusion exponents at each time point. The STEP model is trained using the simulation-based method and validated with experimental data. This includes segmenting the trajectories of the integrin $\alpha5 \beta1$ and pathogen recognition receptor DC-SIGN, both exhibiting varying diffusion coefficients. Notably, STEP demonstrates excellent performance in segmenting these complex trajectories, effectively identifying states with different diffusion coefficients [figure 6(b)].

Another example of point-wise method is the U-AnDi model proposed by X. Qu et al. \cite{Qu2024}. As illustrated in figure 6(c), this model combines dilated causal convolutions to capture long-range dependencies, gated activation units for time-series information filtering, and a U-Net architecture to achieve fine-grained segmentation performance. Similar to the STEP method, U-AnDi predicts diffusion parameters for each point along a trajectory, including the diffusion exponent and diffusion model. As shown in figure 6(d), after training on simulated datasets, U-AnDi is further tested on experimental data involving the diffusion of transmembrane proteins on cell membrane surfaces. These experimental trajectories include two diffusion models: CTRW and FBM, with U-AnDi demonstrating excellent segmentation performance for both. In addition to the STEP and U-AnDi models, other machine learning approaches like RNN \cite{Martinez2023, seckler2024change} and GNN \cite{Pineda2023} also employ point-wise methods to segment trajectories effectively. Detailed information regarding these models is listed in Table 1.

\subsubsection{Further discussions}\label{subsubsec2}

Both the sliding window and the point-wise method are effective in segmenting heterogeneous diffusion dynamics, yet each has its limitations. For the sliding window method, the selection of window length crucially impacts the model's recall in detecting short-term changes. In contrast, the point-wise method benefits from its per-point prediction characteristic, which allows for more effective captures of transient states. However, every point along a trajectory can potentially be identified as a changepoint when employing the point-wise method. This can result in an increased incidence of false positives, reducing the precision of changepoint detection under certain conditions. Therefore, effective post-processing of predictions to filter out false changepoints is crucial for enhancing the segmentation performance of the point-wise method. Potential strategies to mitigate these false positives include the implementation of statistical tests to confirm the significance of potential changepoints, or the application of smoothing algorithms that help discern genuine changepoints from noise. Additionally, extra machine learning models can be trained to learn the typical patterns of changepoints in specific contexts, thereby reducing the likelihood of false detections. Although this post-processing issue has been partially addressed in the studies mentioned \cite{Requena2023,Qu2024}, it remains a topic that requires continued research.

\subsection{Anomalous Diffusion Challenge}\label{subsec2}

Considering the proliferation of machine learning models aimed at analyzing anomalous diffusion, a standardized framework for evaluating their effectiveness remains essential. In the field of machine learning, benchmarking models against established datasets, like ImageNet \cite{Deng2009} for image classification or COCO \cite{Lin2014} for object detection, is the standard practice. Similarly, for the analysis of anomalous diffusion, establishing robust benchmarks is crucial to objectively assess and compare the performance of different models. This has led to the adoption of competitive platforms, similar to Kaggle \cite{kaggle}, where machine learning models contend under uniform conditions using predefined tasks and datasets. In this section, we introduce the Anomalous Diffusion (AnDi) Challenge \cite{andi_website}, which is a machine learning competition specifically designed for the analysis of anomalous diffusion trajectories.

{The AnDi Challenge is organized by a group of researchers \cite{munoz2020anomalous} and is hosted on the CodaLab platform \cite{codalab}.} This competition features two distinct versions. The first iteration in the year 2020 focuses primarily on the inference of diffusion parameters, including the diffusion exponent and diffusion model \cite{Munoz2021}. While this version includes a segmentation task, its simplistic design does not fully replicate the complexities encountered in real-world scenarios. Consequently, the subsequent iteration of the challenge in the year 2024 shifts its emphasis towards the segmentation of heterogeneous diffusion dynamics \cite{Munoz2023}, aiming to address this complex problem more directly and effectively. Details of the AnDi Challenge are listed in Table 2, including the task descriptions, evaluation metrics, number of teams for each task, and links to related works. The ranking information can be found on the official competition website \cite{andi_website} and is not listed in the table here for brevity. In the following, we provide an overview of each version of the AnDi Challenge, focusing on their specific objectives and respective datasets:

\begin{figure}
\centering
\includegraphics[width=0.95\textwidth]{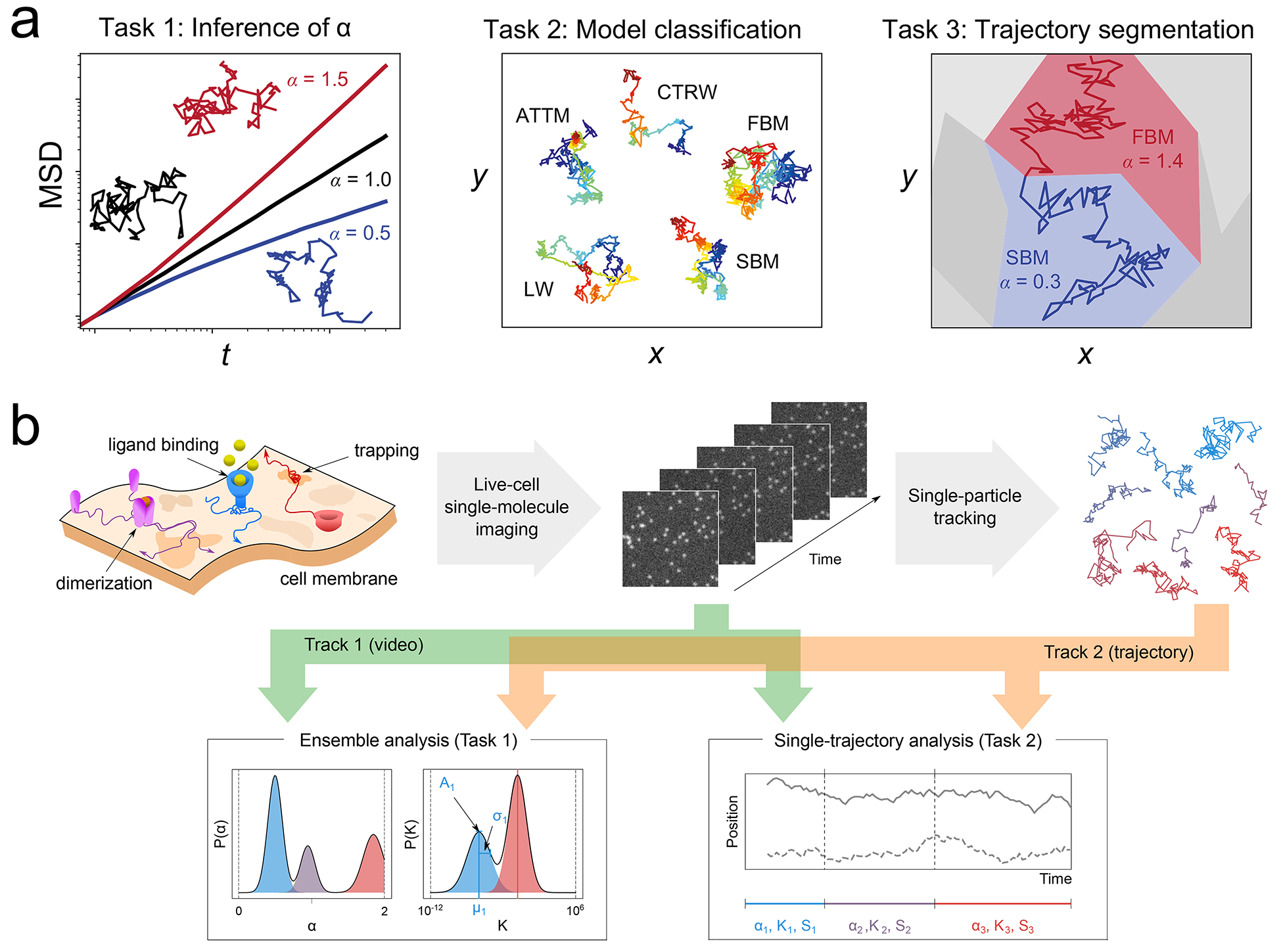}
\caption{(a) Schematic diagram of three tasks in the $1^{\rm st}$ AnDi Challenge, from \cite{Munoz2021}; (b) Illustration of the two competition tracks and corresponding two subtasks in the $2^{\rm nd}$ AnDi Challenge, from \cite{Munoz2023}.}
\label{fig:fig7}
\end{figure}

\begin{sidewaystable}
\caption{Detailed information of the AnDi Challenge. }\label{tab3}
\begin{tabular*}{\textheight}{@{\extracolsep\fill}m{1.3cm}m{8cm}m{3cm}m{3cm}m{2cm}}
\toprule%
{Version} & {Task description} & {Metric} & {Team number} & {Related works} \\
\midrule
$1^{\rm st}$ &
\begin{itemize}[leftmargin=*]
\item Task 1: Inference of the diffusion exponent. Participants are required to infer the diffusion exponent of trajectories.
\item Task 2: Classification of the diffusion model. Participants need to classify the trajectories into five diffusion models: ATTM, CTRW, FBM, LW, and SBM.
\item Task 3: Two-state trajectory segmentation. The goal is to identify the single changepoint in each trajectory where the diffusion state changes just once, and then determine the diffusion model and exponent for each segment.
\end{itemize}
&
\begin{itemize}[leftmargin=*]
\item Task 1: MAE
\item Task 2: F1-Score
\item Task 3: MAE, F1-Score, RMSE
\end{itemize}
&
\begin{itemize}[leftmargin=*]
\item Task 1: 13 teams
\item Task 2: 14 teams
\item Task 3: 4 teams
\end{itemize}
&
\cite{Munoz2021, Manzo2021, Park2021, Krog, Li2021, DeepSPT2020, NOA2020, Argun2021, Garibo-i-Orts2021, Verdier2021, Gentili2021, AnDi-unicorns, ErasmusMC, FCI, Aghion, Wust-ML-A, Wust-ML-B}
\\
\midrule
$2^{\rm nd}$ &
\begin{itemize}[leftmargin=*]
\item This challenge is organized along two tracks:
\item Video track: For analysis of raw SPT videos.
\item Trajectory track: For the analysis of trajectory data.
\end{itemize}
\begin{itemize}[leftmargin=*]
\item Each track has two tasks:
\item Task 1 (Single-trajectory task): For each trajectory, participants are supposed to list the changepoints. After that, they need to determine the diffusion coefficient, diffusion exponent, and motion type for each segment.
\item Task 2 (Ensemble task): For each experimental condition, participants are required to provide the model type, number of states, mean and standard deviation of diffusion coefficients, and diffusion exponents for each state.
\end{itemize}
&
\begin{itemize}[leftmargin=*]
\item Task 1: JSC, RMSE, MSLE, MAE, F1-Score
\item Task 2: Wasserstein distance
\end{itemize}
&
\begin{itemize}[leftmargin=*]
\item Video track:
\begin{itemize}[leftmargin=10pt]
\item Task 1: 3 teams
\item Task 2: 3 teams
\end{itemize}
\item Trajectory track:
\begin{itemize}[leftmargin=10pt]
\item Task 1: 18 teams
\item Task 2: 11 teams
\end{itemize}
\end{itemize}
&
\cite{Munoz2023,seckler2024change}
\\
\bottomrule
\end{tabular*}
\footnotetext{Note: New abbreviations in the table are listed as follows: RMSE: Root mean squared error; JSC: Jaccard similarity coefficient; MSLE: Mean squared logarithmic error.}
\end{sidewaystable}

\begin{enumerate}
\item As shown in figure 7(a), the $1^{\rm st}$ AnDi Challenge contains three tasks: the inference of diffusion exponent (Task 1), classification of diffusion model (Task 2), and segmentation of two-state trajectories (Task 3). Each task is further divided by the space dimensionality (1D, 2D, and 3D), resulting in nine independent subtasks. The trajectory data is generated by the Python package {\it andi\_datasets} \cite{Gorka2020}, based on five mathematical diffusion models: Annealed transient time motion (ATTM) \cite{Massignan}, CTRW \cite{Scher, Dechant2019}, FBM \cite{Mandelbrot}, LW \cite{Klafter2}, and Scaled Brownian motion (SBM) \cite{Lim}. By setting specific parameters in diffusion models, this Python package can generate synthetic diffusion trajectories with known labels. Participants can train their machine learning models using these simulated trajectories. A shared test set is then used by organizers to objectively evaluate the performance of different models, where the results are summarized in Ref. \cite{Munoz2021}. 

\item The $2^{\rm st}$ AnDi Challenge is organized along two tracks: video track and trajectory track. The video track focuses on processing raw SPT videos, while the trajectory track deals with the segmentation of trajectory data. As presented in figure 7(b), each track has two tasks: the single-trajectory task (Task 1), which involves segmenting individual trajectories, and the ensemble task (Task 2), which aims to estimate global parameters used in trajectory simulations. Here, we specifically detail the trajectory track, which is directly relevant to this article. To accurately replicate real experimental conditions of heterogeneous diffusion dynamics, the trajectory track's dataset and objectives have been structured to reflect a higher level of complexity. Thus, to simplify the competition tasks, participants are required to solely process 2D trajectory data. These 2D trajectories are simulated based on five experimental models, namely single-state, multi-state, dimerization, transient-confinement, and quenched-trap models, which can emulate real anomalous diffusion experimental systems. By selecting experimental models and setting parameters, {\it andi\_datasets} can generate a large number of simulated trajectories and corresponding labels. Utilizing these data, participants can train their models to segment trajectories and infer experimental parameters. The final results are summarized in Ref. \cite{Munoz2023} and will be officially published soon.
\end{enumerate}

As subsequent versions continue to be refined, the AnDi Challenge can potentially establish crucial benchmarking protocols for the objective assessment of methods regarding the analysis of anomalous diffusion. Compared to the ImageNet dataset for image classification models, the AnDi Challenge aims to provide a similarly impactful framework for evaluating related models. This challenge facilitates the comparison of diverse methods under uniform testing conditions, which is vital for advancing the field towards robust and generalized solutions.

However, despite its significant contributions, the AnDi Challenge highlights a critical gap in this field: the reluctance within the scientific community to share real experimental data openly. Numerous studies on the machine learning characterization of anomalous diffusion rely heavily on the simulation-based method due to the lack of accessible real datasets. This situation constrains the development of machine learning tools that could be effectively applied to actual experimental analysis. To address this issue, there is a pressing need for the SPT research community to actively promote data sharing. {In particular, establishing standardized sharing protocols is also important to ensure data quality, consistency, and interoperability.} Encouraging the open source of real experimental data {under these protocols} would not only validate the applicability of these models in practical settings, but also accelerate the discovery of novel insights into the dynamics of anomalous diffusion processes.

\section{Representation Learning of Anomalous Diffusion}

Moving beyond the application-focused tasks of characterizing anomalous diffusion trajectories, a deeper understanding of representation learning, i.e., how to effectively represent the anomalous diffusion, emerges as a critical foundational step in the analysis of such complex phenomena. Effective representation of anomalous diffusion is crucial not only for enabling practical tasks like classification, segmentation, and parameter estimation, but also for embedding the intricate diffusion dynamics into a comprehensible and operational framework. The diversity and complexity of physical mechanisms underlying different types of anomalous diffusion necessitate versatile representation strategies. A robust representation of trajectory data, therefore, should distill the essence of corresponding diffusion dynamics into a form that is informative, discriminative, and adaptable to various analytical challenges.

To address this issue, we introduce three distinct strategies for representing the characteristics of anomalous diffusion in this section. The first involves the combination of predefined and handcrafted features to encapsulate the essential physical information of diffusion processes. The second employs neural networks to derive feature vectors from the penultimate layer, capturing a data-driven representation of diffusion dynamics. Lastly, we discuss the strategy of using latent representations through autoencoder (encoder-decoder) architectures \cite{Zhai2018}, which not only reflect the underlying mechanisms but also facilitate the reconstruction of trajectories.

\subsection{Diffusion fingerprint: Combination of predefined features}

Prior to the widespread integration of machine learning into the study of anomalous diffusion, researchers have already employed a range of physical parameters to characterize the diffusion dynamics, such as MSD and diffusion exponent. In contrast to machine learning methods that autonomously extract features from data, the derivation of these parametric features necessitates a deep understanding of the physical principles underlying anomalous diffusion. Consequently, they are typically predefined and manually extracted, anchored firmly in domain-specific knowledge. Therefore, the combination of these predefined features not only captures the complex behaviors inherent in anomalous diffusion but also provides substantial interpretability.

\begin{table}[h]
\raggedleft
\caption{Overview of predefined feature combinations used in representative studies to characterize anomalous diffusion through feature-based models. In Ref. \cite{Loch2020}, the authors evaluate the performance of three different feature combinations separately, designated as Set A, Set B, and Set C. }
\begin{tabular*}{\textwidth}{@{\extracolsep\fill}lcccccc}
\toprule%
  ~ & \multicolumn{3}{@{}c@{}}{Random forest \cite{Loch2020}}  & LR \cite{Pinholt2021} & XGBoost \cite{Kowalek2022}\\
    ~ & Set A & Set B & Set C & & \\
  \midrule
    Anomalous exponent  &\checkmark&\checkmark&\checkmark&\checkmark& \checkmark\\
    Diffusion coefficient  &\checkmark&\checkmark&\checkmark&\checkmark& \checkmark\\
    MSD ratio &\checkmark& \checkmark& &\checkmark& \checkmark\\
    Efficiency  &\checkmark&\checkmark& &\checkmark& \checkmark \\
    Straightness &\checkmark&\checkmark& &&\checkmark\\
    Empirical VACF &\checkmark& & && \checkmark \\
    Maximal excursion &\checkmark& & && \checkmark\\
    {\it p}-variation-based statistics &\checkmark& & & &\checkmark \\
    Asymmetry & &\checkmark& &&\checkmark  \\
    Fractal dimension & &\checkmark& &\checkmark& \checkmark \\
    Gaussianity & &\checkmark& &\checkmark& \checkmark   \\
    Kurtosis & &\checkmark& &\checkmark&\checkmark \\
    Trappedness & &\checkmark& &\checkmark&\checkmark \\
    Standardised maximum distance & & &\checkmark& &   \\
    Exponent of power function & & &\checkmark&&      \\
    Residence times in each state &&&&\checkmark& \\
    Average residence time &&&&\checkmark& \\
    Average step length &&&&\checkmark& \\
    Quality of power law fit  &&&&\checkmark& \\
    Intermediate time spread &&&&\checkmark& \\
    Number of points in a trajectory &&&&\checkmark& \\
    Mean maximal excursion & & & &&\checkmark \\
    D{’}Agostino-Pearson test statistic & & & &&\checkmark \\
    Kolmogorov-Smirnov statistic  & & & && \checkmark \\ 
    Noah exponent & & & &&  \checkmark\\
    Moses exponent & & & && \checkmark\\
    Joseph exponent & & & && \checkmark \\
    Detrending moving average & & & && \checkmark \\
    Average moving window characteristics & & & && \checkmark \\
    Maximum standard deviation & & & &&\checkmark  \\
\botrule
\end{tabular*}
\footnotetext{Note: Here, LR is the abbreviation of logistic regression.}
\end{table}

As discussed in section 2.1.1, feature-based machine learning models successfully leverage these predefined features to infer diffusion parameters. Nevertheless, selecting the most effective combination of predefined features to represent anomalous diffusion accurately remains a central challenge. Extensive research employing feature-based models has explored a diverse array of feature combinations to effectively characterize the diffusion dynamics. For an intuitive presentation, we summarize the feature combinations used in three representative works \cite{Loch2020, Pinholt2021, Kowalek2022} in Table 3. As we see in this table, the selected features encompass not only specific anomalous diffusion parameters such as the diffusion exponent and MSD but also standard statistical measures like kurtosis and standard deviation. This composition of predefined features is strategically crafted to represent diffusion trajectories within an established knowledge framework. It maximizes the informational content while ensuring that the features maintain clear physical meanings and robust interpretability.

\begin{figure}
\centering
\includegraphics[width=0.95\textwidth]{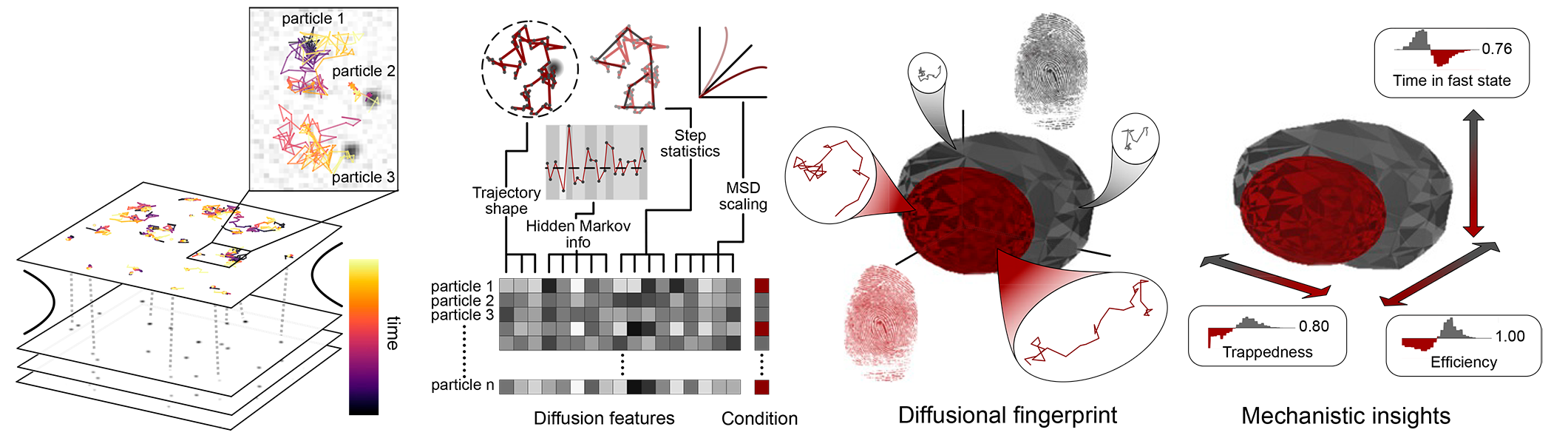}
\caption{Schematic diagram of the diffusion fingerprint for representing and analyzing the single-particle tracking trajectory data, from \cite{Pinholt2021}.}
\label{fig:fig8}
\end{figure}

Building upon this framework, as depicted in figure 8, H. D. Pinholt et al. \cite{Pinholt2021} advance the field of representation learning by introducing the term ``diffusion fingerprint" to describe the unique characteristics of combined features. This innovative concept underscores the precision with which these feature combinations can delineate the dynamic profiles of different diffusion types, serving a role analogous to how a fingerprint distinctively identifies an individual. However, as highlighted in section 2.1.1, the study by H. Loch-Olszewska et al. \cite{Loch2020} emphasizes that no single feature combination excels across all types of anomalous diffusion. This observation suggests that there are inherent limitations in the representational power of predefined feature combinations when characterizing the properties of anomalous diffusion.

Therefore, to enhance the representational capacity of feature combinations while maintaining their high interpretability, further exploration of new and effective diffusion or statistical features beyond those already known is necessary. For example, M. Mangalam et al. \cite{Mangalam2023} have advanced the analysis of non-ergodic diffusion processes by incorporating fractal and multifractal descriptors that capture ergodic properties. By employing the Chhabra-Jensen method \cite{Chhabra1989}, they calculate the multifractal spectrum curve, which relates the singularity exponents of these descriptors to the Hausdorff dimension. This approach effectively illustrates the nonlinearity and complexity inherent in different types of diffusion. For instance, the multifractal spectrum of FBM typically appears symmetric, reflecting relatively consistent fluctuations throughout the process. Conversely, SBM exhibits an asymmetric singularity spectrum, indicating fluctuations concentrated at particular scales. Leveraging the discriminative capabilities of the multifractal spectrum, H. Seckler et al. \cite{Seckler2024} further select nine spectrum-related parameters to form a multifractal spectral (MFS) feature set, which serves as a novel representation for diffusion processes. By integrating these MFS features with the predefined features listed in the XGBoost column of Table 3 , the classification performance of machine learning models is substantially enhanced. This integration underscores the effectiveness of MFS features in enriching the diffusion feature landscape, highlighting the importance of mining new features for the analysis of anomalous diffusion.

\subsection{Data-driven representation learning by neural networks}

The existing frameworks of knowledge do not completely capture all the nuanced features of diffusion processes, inherently imposing an upper limit on feature discovery based on traditional understandings. This limitation becomes glaringly apparent in efforts to characterize complex diffusion trajectories. However, inspired by data-driven machine learning approaches, enabling models to autonomously learn from data to extract effective representations could transcend these traditional constraints. This shift would free the analysis from the constraints of existing theoretical frameworks, allowing for the capture of high-level abstract representations of diffusion dynamics that are more effective and encompassing.

This data-driven representation learning, extensively applied in fields such as image, text, and speech analysis, involves transforming raw data into embedded vectors via deep neural networks \cite{ Bengio2013, Zhong2016}. By constructing databases of these embeddings, the approach significantly enhances the efficiency of data retrieval, matching, and classification tasks. Therefore, the representation of anomalous diffusion can also benefit from this data-driven approach, wherein deep neural networks autonomously perform representation learning directly from trajectory data. Here, we introduce two commonly utilized data-driven representations of diffusion processes: feature vector from the penultimate layer of neural network, and latent representation from the autoencoder.

\subsubsection{Feature vector from the penultimate layer of neural network}

In general, when employing a neural network for parameter inference, the final prediction is determined by the linear operations in the last fully connected layer. In particular, these simple operations are only applied to the output from the penultimate layer, indicating that the model performance is substantially dependent on the representational power of this output. Similarly, in high-performing neural networks tailored for anomalous diffusion analysis, the feature vector from the penultimate layer encapsulates significant information that delineates the diffusion dynamics. This comprehensive content is vital for the model's robust performance, making this feature vector a potent representation of the diffusion process.

\begin{figure}
\centering
\includegraphics[width=0.95\textwidth]{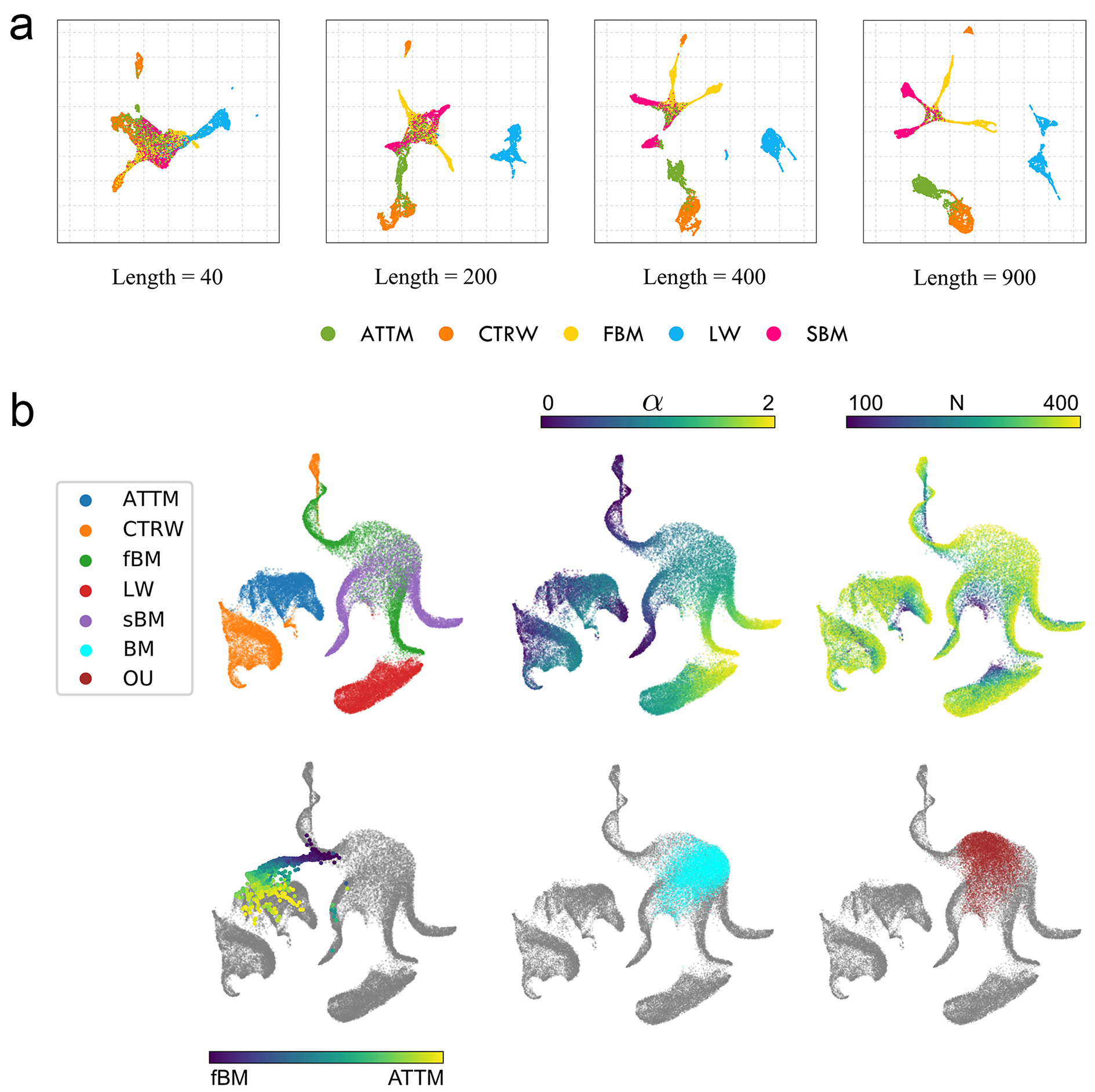}
\caption{(a) UMAP visualization of feature vectors of diffusion trajectories from WADNet across five different diffusion models and varying lengths, from \cite{Li2021}; (b) UMAP representation of trajectory feature vectors from the GNN model. Each plot, arranged from left to right and from top to bottom, focuses on the following aspects respectively: diffusion model, diffusion exponent, trajectory length, mixed trajectories, BM, and OU process, from \cite{Verdier2021}.}
\label{fig:fig9}
\end{figure}

However, representations derived from data-driven methods typically lack explicit physical or statistical meaning, rendering them less interpretable. Despite this limitation, it is still possible to gain insights into how these feature vectors represent anomalous diffusion characteristics by examining their distributions through visualization techniques. Dimensionality reduction techniques like t-distributed stochastic neighbor embedding (t-SNE) \cite{VanderMaaten2008} and uniform manifold approximation and projection (UMAP) \cite{McInnes2018} allow for the visualization of high-dimensional feature vectors in a lower-dimensional space. For example, D. Li et al. \cite{Li2021} utilize UMAP to visualize feature vectors from WADNet on a 2D plane, as illustrated in figure 9(a). This visualization specifically represents feature vectors of trajectories across five different diffusion models and varying lengths. It is evident that as the trajectory length increases, trajectories from different diffusion models become more distinctly separated. This result demonstrates that the representational ability of feature vectors to differentiate among these diffusion models becomes more pronounced with increasing trajectory lengths. In addition, the cluster distribution distinctly separates the motion patterns of LW from the others, while FBM and SBM show overlapping behaviors, as do ATTM and CTRW. This distribution align with the theoretical definitions of these diffusion models, indicating that the feature vectors from the penultimate layer of WADNet effectively capture the characteristics of diffusion trajectories.

Another work that highlights the feature vectors from neural networks is by H. Verdier et al. \cite{Verdier2021}, where these vectors are extracted by a GNN model. As illustrated in figure 9(b), the UMAP visualizations capture the diffusion dynamics across various parameters such as diffusion model, diffusion exponent, and trajectory length. The clustering clearly demonstrates the GNN's ability to discern differences among these parameters, underscoring the effectiveness of feature vectors in capturing the essence of diffusion processes. Additionally, the authors assess the robustness and generalizability of these representations by testing mixed trajectories and unseen diffusion models. For mixed trajectories from FBM and ATTM models, the feature vectors exhibit a smooth transition across different types of motion as their proportions vary, effectively interpolating between the characteristics of these two diffusion models. Further, feature vectors from trajectories generated using diffusion models not included in the training set are incorporated, specifically Brownian motion (BM) and the Ornstein-Uhlenbeck (OU) process \cite{Crispin2009}. Corresponding distribution of clusters show that these trajectories overlap significantly with those of SBM and FBM. This consistency with theoretical expectations confirms the capacity of feature vectors from the penultimate layer of neural network to generalize and robustly represent diverse diffusion dynamics.

\subsubsection{Latent representation from the autoencoder}

\begin{figure}
\centering
\includegraphics[width=0.95\textwidth]{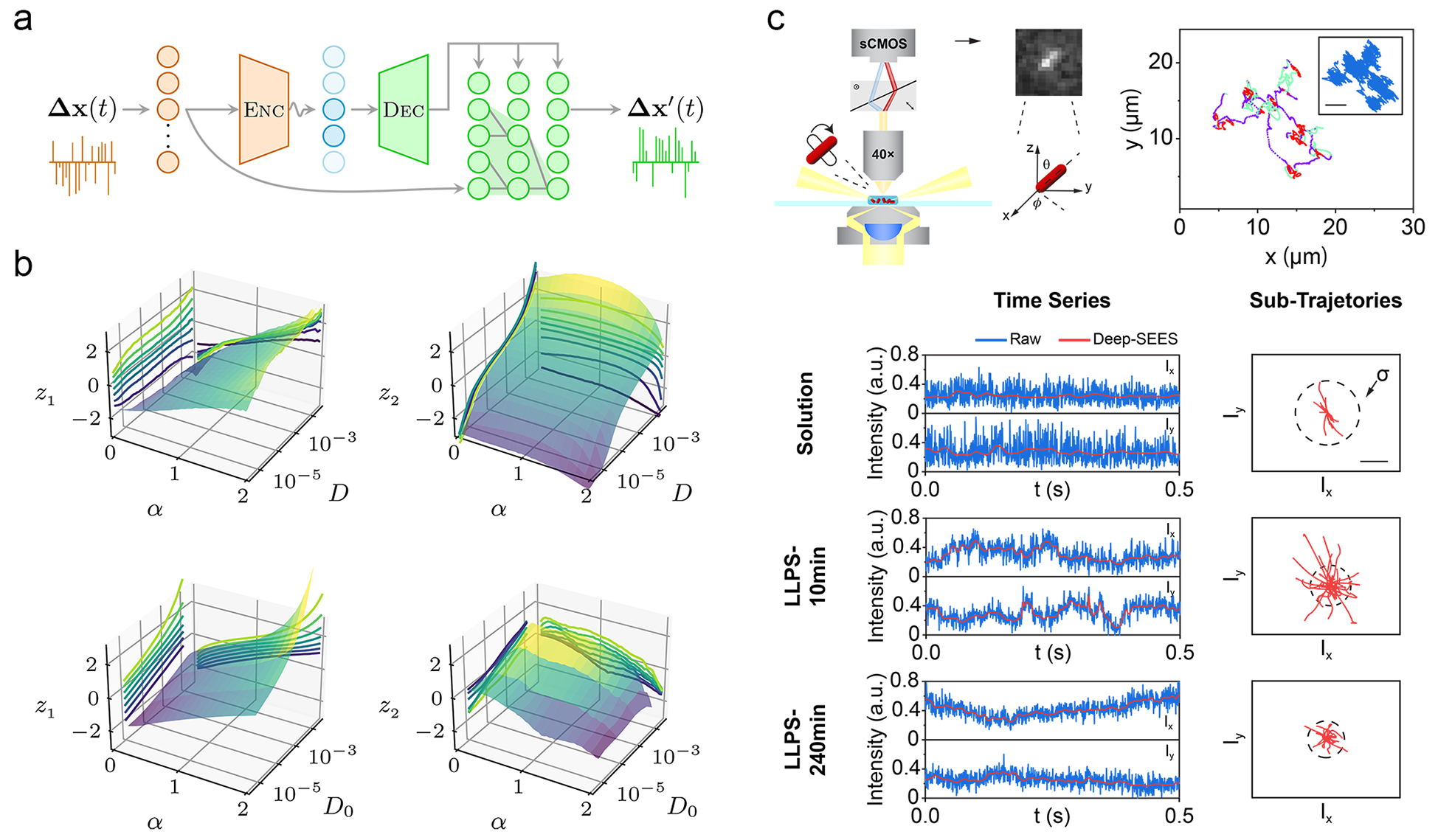}
\caption{(a) Schematic diagram of the autoencoder architecture, from \cite{Fernandez2024}; (b) Distribution of latent neurons activations, showing clear relations to the diffusion parameters, from \cite{Fernandez2024}; (c) Illustration of the experimental system of gold nanorods, where both translational (top) and rotational (bottom) trajectories are effectively denoised by the LSTM-VAE model, from \cite{Zhang2023}.}
\label{fig:fig10}
\end{figure}

While leveraging trained neural networks for parameter inference is a highly effective and efficient method for extracting representations of diffusion trajectories, it is important to note that these models are specifically designed for inference tasks. This specialized focus may inadvertently lead to the neglect of certain trajectory features that do not directly influence the target parameters. As a result, the derived representations can exhibit inherent biases, capturing a skewed perspective of diffusion dynamics. Therefore, developing neural networks that focus primarily on the extraction of representations is crucial for advancing the representation learning of anomalous diffusion.

For that purpose, researchers have shifted their emphasis towards machine learning models based on the autoencoder architecture. This architecture is particularly adept at distilling high-dimensional data into efficient latent representations. As show in figure 10(a), the autoencoder consists of two components: an encoder and a decoder. The encoder compresses the input data into a low-dimensional hidden layer, which forms the latent representation. The decoder then attempts to reconstruct the original input data from this latent space. It is evident that the training of autoencoder models does not rely on data labels, making them considerably more suitable for representation learning than discriminative models that require labeled data.

The pioneering work in this area is conducted by G. Mu\~{n}oz-Gil et al. \cite{MunozUnsupervised}, where an autoencoder model is utilized for anomaly detection in an unsupervised way. Although this work does not specifically highlight the representation learning of anomalous diffusion, their application of autoencoder and realization of trajectory reconstruction has significantly inspired subsequent research in this domain. Following this work, G. Fern\'{a}ndez-Fern\'{a}ndez et al. \cite{Fernandez2024} introduce a $\beta$-variational autoencoder ($\beta$-VAE) tailored for the representation learning of stochastic processes. {Instead of explicitly fixing the number of latent neurons, they use an annealing schedule for the parameter $\beta$, which ensures that only the minimal number of informative latent neurons remain. This allows the model to focus on capturing the essential physical parameters, such as the diffusion coefficient and the diffusion exponent, while still accurately reconstructing the underlying trajectories. Consequently, the latent neurons that remain informative after this regularization process exhibit a strong correlation with these key physical variables [figure 10(b)]. This correlation is not a mere by-product of dimensionality reduction but results from the balance between reconstruction quality and the $\beta$-weighted Kullback-Leibler (KL) divergence. As a result, the approach enhances the interpretability of the learned representations and enables the autonomous discovery of previously unknown parameters.}

On the other hand, autoencoders excel in processing experimental trajectories directly due to their independence from labeled data. This attribute enables versatile applications in managing raw data, particularly effective in denoising experimental trajectories through latent representation and trajectory reconstruction. For example, the Deep-SEES method, developed by Y. Zhang et al. \cite{Zhang2023}, utilizes the LSTM-VAE model to extract latent representations from experimental trajectories. This approach enables the reconstruction of these trajectories, effectively denoising and preserving their physical characteristics. The robustness of these reconstructions is tested across challenging experimental conditions, including trajectories of gold nanorods in active enzyme systems and protein liquid-liquid phase separation systems. These trajectories are characterized by high noise levels due to low photon flux during data acquisition. As highlighted in figure 10(c), the LSTM-VAE model effectively denoises these trajectories for both translational and rotational movements, maintaining their physical accuracy and facilitating the subsequent classification of different motion types. The results not only demonstrate the exceptional performance of autoencoders in denoising but also affirms the potent representational capability of latent representations in capturing the dynamics of diffusion processes.

\subsection{Further discussions}

\begin{table}[h]
\raggedleft
\caption{Comparison of three methods for the representation learning of anomalous diffusion. }
\begin{tabular*}{\textwidth}{@{\extracolsep\fill}lccc}
  \toprule
  ~ & Predefined feature & Feature vector & Latent\\
  \midrule
    Base of representation & Known knowledge & Data & Data\\
    Representational power & Moderate & Strong & Strong \\
    Interpretability & Strong & Weak & Weak \\
    Dependence on labels & Weak & Strong & Weak \\
    Computational demand  & Low  & Moderate & High \\
    Automatic feature discovery & Incapable & Able & Able \\
    Ability to reconstruct & Incapable  & Incapable  & Able  \\

\botrule
\end{tabular*}
\footnotetext{Note: Here, ``predefined feature" refers to the combination of predefined features, ``feature vector" to the feature vector from the penultimate layer of neural network, and ``latent" to the latent representation from the autoencoder.}
\end{table}
As detailed in the preceding sections, each of the three methods for the representation learning of anomalous diffusion specializes in distinct aspects, highlighting their unique contributions to understanding diffusion processes. For applications where interpretability is paramount, such as in scientific research where understanding the mechanisms of diffusion is crucial, predefined features might be preferred. In contrast, for applications where model performance and adaptability to complex data are more critical, feature vectors from neural networks and latent representations from autoencoders may offer superior solutions. To provide a clearer overview, we have compared the three methods across multiple dimensions in Table 4, offering valuable insights into choosing the most appropriate representation technique for analyzing diffusion processes. Future research may consider the use of hybrid models that leverage both data-driven features and domain-specific knowledge. This strategy could harness the collective strengths of these two types of representations, potentially establishing enhanced benchmarks for the analysis of anomalous diffusion.

\section{Conclusions}

In summary, the integration of machine learning techniques in the analysis of anomalous diffusion is systematically introduced in this paper. We have discussed several key areas in this field, including the single trajectory characterization via machine learning and the representation learning of anomalous diffusion. For the characterization of single trajectory properties, machine learning techniques demonstrate superior performance compared to traditional methods in inferring diffusion parameters and segmenting trajectories. Moreover, the variety of machine learning approaches, spanning from classical machine learning to advanced deep learning, provides an extensive array of options that can be customized to meet the distinct demands of different scenarios. Through machine learning competitions such as the AnDi Challenge, the field has pushed towards standardized benchmarks that facilitate the evaluation and comparison of various analytical models under uniform conditions.

Regarding the representation learning of anomalous diffusion, we introduce and compare three principal strategies: the combination of predefined features, the feature vector from the penultimate layer of neural network, and the latent representation from the autoencoder. Each strategy has its merits and limitations, suggesting that the choice of method should be aligned with specific research needs. Predefined features, while limited in adaptability, offer significant interpretability, making them invaluable in contexts where understanding the underlying physical mechanisms is paramount. On the other hand, data-driven representations, including feature vectors and latent representations, provide powerful tools for handling more complex trajectories, although they often lack the actual physical meanings.

Despite substantial advances, this field still faces significant challenges, particularly in terms of data availability, encompassing both simulated and experimental datasets. Currently, the simulated data available generally do not fully replicate the complexities of real-world trajectories, with many diffusion models lacking corresponding simulation tools. This restriction often confines research to diffusion models included in available tools such as {\it andi-datasets} \cite{andi_dateset}. Thus, increasing the availability of trajectory simulation software is essential, which would significantly enhance the capabilities of machine learning models trained via simulation-based methods. On the other hand, the access to experimental trajectory data is largely restricted, which hinders the involvement of machine learning experts in anomalous diffusion research. Therefore, the establishment of an open-source ecosystem for sharing experimental data is important, requiring collective efforts from the research community to accelerate development in this domain.

Moving forward, machine learning analysis of anomalous diffusion presents two promising research directions. The first involves the creation of innovative methods and models to address complex analytical challenges in trajectory analysis, such as accurate changepoint detection and forecasting of random walks. This direction necessitates a focus on advanced techniques for time-series analysis, potentially leveraging cutting-edge technologies like large language models \cite{Gruver2024}. The second emphasizes enhancing the interpretability of machine learning models to facilitate a deeper understanding of the physical underpinnings of diffusion processes. This will require embedding domain-specific knowledge of anomalous diffusion into machine learning architectures to enable more insightful analysis and interpretation. In conclusion, leveraging machine learning to investigate anomalous diffusion represents a vital approach that is reshaping our understanding of complex systems. This progressive integration not only promises to refine our predictive capabilities but also holds potential to broaden our theoretical frameworks of anomalous diffusion, thereby offering profound insights into the complex behaviors and interactions that drive diffusion phenomena across physical, biological, and technological domains.

\bmhead{Acknowledgements}

We thank Yu Zhang for helpful discussions. This work is supported by the National Natural Science Foundation of China (Grant No. 12104147) and the Fundamental Research Funds for the Central Universities.

\bmhead{Data Availability Statement}

No data associated in the manuscript. One can obtain the relevant datasets and materials from the references below.

\bibliography{sn-bibliography}

\begin{thebibliography}{100}
\providecommand{\url}[1]{{#1}}
\providecommand{\urlprefix}{URL }
\providecommand{\doi}[1]{\url{https://doi.org/#1}}
\bibcommenthead

\bibitem{Metzler2014}
R.~Metzler, J.H. Jeon, A.G. Cherstvy, E.~Barkai, Anomalous diffusion models and
  their properties: non-stationarity, non-ergodicity, and ageing at the
  centenary of single particle tracking.
\newblock Phys. Chem. Chem. Phys. \textbf{16}(44), 24128--24164 (2014)

\bibitem{Klafter2015}
J.~Klafter, I.M. Sokolov, Anomalous diffusion spreads its wings.
\newblock Phys. World \textbf{18}(8), 29 (2005)

\bibitem{Manzo2023preface}
C.~Manzo, G.~Mu{\~n}oz-Gil, G.~Volpe, M.A. Garcia-March, M.~Lewenstein,
  R.~Metzler, Preface: characterisation of physical processes from anomalous
  diffusion data.
\newblock J. Phys. A: Math. Theor. \textbf{56}(1), 010401 (2023)

\bibitem{Mason1995}
T.G. Mason, D.A. Weitz, Optical measurements of frequency-dependent linear
  viscoelastic moduli of complex fluids.
\newblock Phys. Rev. Lett. \textbf{74}(7), 1250 (1995)

\bibitem{Aarao2016}
F.D.A. Aar{\~a}o~Reis, Scaling relations in the diffusive infiltration in
  fractals.
\newblock Phys. Rev. E \textbf{94}(5), 052124 (2016)

\bibitem{Volpe2017}
G.~Volpe, G.~Volpe, The topography of the environment alters the optimal search
  strategy for active particles.
\newblock Proc. Natl Acad. Sci. USA \textbf{114}(43), 11350--11355 (2017)

\bibitem{Barbosa2024}
S.~Barbosa, M.~Kiefer-Emmanouilidis, F.~Lang, J.~Koch, A.~Widera,
  Characterizing localization effects in an ultracold disordered {F}ermi gas by
  diffusion analysis.
\newblock Phys. Rev. Res. \textbf{6}(3), 033039 (2024)

\bibitem{Iida2024}
K.~Iida, A.~Dechant, T.~Akimoto, Universality of giant diffusion in tilted
  periodic potentials.
\newblock arXiv:2404.12761  (2024)

\bibitem{Hegde2022}
A.S. Hegde, C.M. Chandrashekar, Characterization of anomalous diffusion in
  one-dimensional quantum walks.
\newblock J. Phys. A: Math. Theor. \textbf{55}(23), 234006 (2022)

\bibitem{Vitali2022}
S.~Vitali, P.~Paradisi, G.~Pagnini, Anomalous diffusion originated by two
  {M}arkovian hopping-trap mechanisms.
\newblock J. Phys. A: Math. Theor. \textbf{55}(22), 224012 (2022)

\bibitem{Shi2023}
H.~Shi, L.~Du, F.~Huang, W.~Guo, Weak ergodicity breaking and anomalous
  diffusion in collective motion of active particles under spatiotemporal
  disorder.
\newblock Phys. Rev. E \textbf{107}(2), 024114 (2023)

\bibitem{Zheng2022}
C.~Zheng, R.~T{\"o}njes, Noise-induced swarming of active particles.
\newblock Phys. Rev. E \textbf{106}(6), 064601 (2022)

\bibitem{Xu2021}
Z.~Xu, X.~Dai, X.~Bu, Y.~Yang, X.~Zhang, X.~Man, X.~Zhang, M.~Doi, L.T. Yan,
  Enhanced heterogeneous diffusion of nanoparticles in semiflexible networks.
\newblock ACS Nano \textbf{15}(3), 4608--4616 (2021)

\bibitem{Dai2022}
X.~Dai, X.~Zhang, L.~Gao, Z.~Xu, L.T. Yan, Topology mediates transport of
  nanoparticles in macromolecular networks.
\newblock Nat. Commun. \textbf{13}(1), 4094 (2022)

\bibitem{Muller1992}
F.~M{\"u}ller-Plathe, S.C. Rogers, W.F. van Gunsteren, Computational evidence
  for anomalous diffusion of small molecules in amorphous polymers.
\newblock Chem. Phys. Lett. \textbf{199}(3-4), 237--243 (1992)

\bibitem{Ding2014}
Y.~Ding, A.A. Hassanali, M.~Parrinello, Anomalous water diffusion in salt
  solutions.
\newblock Proc. Natl Acad. Sci. USA \textbf{111}(9), 3310--3315 (2014)

\bibitem{Barkai2012}
E.~Barkai, Y.~Garini, R.~Metzler, Strange kinetics of single molecules in
  living cells.
\newblock Phys. Today \textbf{65}(8), 29--35 (2012)

\bibitem{Wu2000}
X.L. Wu, A.~Libchaber, Particle diffusion in a quasi-two-dimensional bacterial
  bath.
\newblock Phys. Rev. Lett. \textbf{84}(13), 3017 (2000)

\bibitem{Illukkumbura2020}
R.~Illukkumbura, T.~Bland, N.W. Goehring, Patterning and polarization of cells
  by intracellular flows.
\newblock Curr. Opin. Cell Biol. \textbf{62}, 123--134 (2020)

\bibitem{Gonzalez2008}
M.C. Gonz\'{a}lez, C.A. Hidalgo, A.L. Barab\'{a}si, Understanding individual
  human mobility patterns.
\newblock Nature \textbf{453}(7196), 779--782 (2008)

\bibitem{Wang2013}
B.~Wang, J.~Kuo, S.~Granick, Bursts of active transport in living cells.
\newblock Phys. Rev. Lett. \textbf{111}(20), 208102 (2013)

\bibitem{Chen2016bio}
P.~Chen, Z.~Huang, J.~Liang, T.~Cui, X.~Zhang, B.~Miao, L.T. Yan, Diffusion and
  directionality of charged nanoparticles on lipid bilayer membrane.
\newblock ACS Nano \textbf{10}(12), 11541--11547 (2016)

\bibitem{Hofling2013}
F.~H{\"o}fling, T.~Franosch, Anomalous transport in the crowded world of
  biological cells.
\newblock Rep. Prog. Phys. \textbf{76}(4), 046602 (2013)

\bibitem{Zhang2021}
M.L. Zhang, H.Y. Ti, P.Y. Wang, H.~Li, Intracellular transport dynamics
  revealed by single-particle tracking.
\newblock Biophys. Rep. \textbf{7}(5), 413 (2021)

\bibitem{Plerou2000}
V.~Plerou, P.~Gopikrishnan, L.A.N. Amaral, X.~Gabaix, H.E. Stanley, Economic
  fluctuations and anomalous diffusion.
\newblock Phys. Rev. E \textbf{62}(3), R3023 (2000)

\bibitem{Masoliver2003}
J.~Masoliver, M.~Montero, G.H. Weiss, Continuous-time random-walk model for
  financial distributions.
\newblock Phys. Rev. E \textbf{67}(2), 021112 (2003)

\bibitem{Jiang2019}
Z.Q. Jiang, W.J. Xie, W.X. Zhou, D.~Sornette, Multifractal analysis of
  financial markets: a review.
\newblock Rep. Prog. Phys. \textbf{82}(12), 125901 (2019)

\bibitem{Meyer2023}
P.G. Meyer, A.G. Cherstvy, H.~Seckler, R.~Hering, N.~Blaum, F.~Jeltsch,
  R.~Metzler, Directedeness, correlations, and daily cycles in springbok
  motion: from data via stochastic models to movement prediction.
\newblock Phys. Rev. Res. \textbf{5}(4), 043129 (2023)

\bibitem{Munoz2021}
G.~Mu{\~n}oz-Gil, G.~Volpe, M.A. Garcia-March, E.~Aghion, A.~Argun, C.B. Hong,
  T.~Bland, S.~Bo, J.A. Conejero, N.~Firbas~{et al.}, Objective comparison of
  methods to decode anomalous diffusion.
\newblock Nat. Commun. \textbf{12}(1), 6253 (2021)

\bibitem{Sposini2022}
V.~Sposini, D.~Krapf, E.~Marinari, R.~Sunyer, F.~Ritort, F.~Taheri,
  C.~Selhuber-Unkel, R.~Benelli, M.~Weiss, R.~Metzler, G.~Oshanin, Towards a
  robust criterion of anomalous diffusion.
\newblock Commun. Phys. \textbf{5}(1), 305 (2022)

\bibitem{Vilk2022sys}
O.~Vilk, E.~Aghion, T.~Avgar, C.~Beta, O.~Nagel, A.~Sabri, R.~Sarfati, D.K.
  Schwartz, M.~Weiss, D.~Krapf~{et al.}, Unravelling the origins of anomalous
  diffusion: from molecules to migrating storks.
\newblock Phys. Rev. Res. \textbf{4}(3), 033055 (2022)

\bibitem{Wang2022}
W.~Wang, A.G. Cherstvy, R.~Metzler, I.M. Sokolov, Restoring ergodicity of
  stochastically reset anomalous-diffusion processes.
\newblock Phys. Rev. Res. \textbf{4}(1), 013161 (2022)

\bibitem{Timashev2010}
S.F. Timashev, Y.S. Polyakov, P.I. Misurkin, S.G. Lakeev, Anomalous diffusion
  as a stochastic component in the dynamics of complex processes.
\newblock Phys. Rev. E \textbf{81}(4), 041128 (2010)

\bibitem{Chen2013}
K.~Chen, B.~Wang, J.~Guan, S.~Granick, Diagnosing heterogeneous dynamics in
  single-molecule/particle trajectories with multiscale wavelets.
\newblock Acs Nano \textbf{7}(10), 8634--8644 (2013)

\bibitem{Bowang2009}
B.~Wang, S.M. Anthony, S.C. Bae, S.~Granick, Anomalous yet {B}rownian.
\newblock Proc. Natl Acad. Sci. USA \textbf{106}(36), 15160--15164 (2009)

\bibitem{Mykyta2014}
M.V. Chubynsky, G.W. Slater, Diffusing diffusivity: a model for anomalous, yet
  {B}rownian, diffusion.
\newblock Phys. Rev. Lett. \textbf{113}(9), 098302 (2014)

\bibitem{Haroldo2023}
H.V. Ribeiro, A.A. Tateishi, E.K. Lenzi, R.L. Magin, M.~Perc, Interplay between
  particle trapping and heterogeneity in anomalous diffusion.
\newblock Commun. Phys. \textbf{6}(1), 244 (2023)

\bibitem{Haroldo2014}
H.V. Ribeiro, A.A. Tateishi, L.G. Alves, R.S. Zola, E.K. Lenzi, Investigating
  the interplay between mechanisms of anomalous diffusion via fractional
  {B}rownian walks on a comb-like structure.
\newblock New J. Phys. \textbf{16}(9), 093050 (2014)

\bibitem{Yasmine2011}
Y.~Meroz, I.M. Sokolov, J.~Klafter, Unequal twins: probability distributions do
  not determine everything.
\newblock Phys. Rev. Lett. \textbf{107}(26), 260601 (2011)

\bibitem{BoWang2012}
B.~Wang, J.~Kuo, S.C. Bae, S.~Granick, When {B}rownian diffusion is not
  {G}aussian.
\newblock Nat. Mater. \textbf{11}(6), 481--485 (2012)

\bibitem{Scher}
H.~Scher, E.W. Montroll, Anomalous transit-time dispersion in amorphous solids.
\newblock Phys. Rev. B \textbf{12}(6), 2455 (1975)

\bibitem{Dechant2019}
A.~Dechant, F.~Kindermann, A.~Widera, E.~Lutz, Continuous-time random walk for
  a particle in a periodic potential.
\newblock Phys. Rev. Lett. \textbf{123}(7), 070602 (2019)

\bibitem{Mandelbrot}
B.B. Mandelbrot, J.W. Van~Ness, Fractional {B}rownian motions, fractional
  noises and applications.
\newblock SIAM Rev. \textbf{10}(4), 422--437 (1968)

\bibitem{Chen2015}
K.~Chen, B.~Wang, S.~Granick, Memoryless self-reinforcing directionality in
  endosomal active transport within living cells.
\newblock Nat. Mater. \textbf{14}(6), 589--593 (2015)

\bibitem{Persson2013}
F.~Persson, M.~Lind{\'e}n, C.~Unoson, J.~Elf, Extracting intracellular
  diffusive states and transition rates from single-molecule tracking data.
\newblock Nat. Meth. \textbf{10}(3), 265--269 (2013)

\bibitem{Monnier2015}
N.~Monnier, Z.~Barry, H.Y. Park, K.C. Su, Z.~Katz, B.P. English, A.~Dey,
  K.~Pan, I.M. Cheeseman, R.H. Singer, M.~Bathe, Inferring transient particle
  transport dynamics in live cells.
\newblock Nat. Meth. \textbf{12}(9), 838--840 (2015)

\bibitem{Johnson1992}
A.R. Johnson, J.A. Wiens, B.T. Milne, T.O. Crist, Animal movements and
  population dynamics in heterogeneous landscapes.
\newblock Landsc. Ecol. \textbf{7}, 63--75 (1992)

\bibitem{Cicerone1995}
M.T. Cicerone, F.R. Blackburn, M.D. Ediger, Anomalous diffusion of probe
  molecules in polystyrene: evidence for spatially heterogeneous segmental
  dynamics.
\newblock Macromolecules \textbf{28}(24), 8224--8232 (1995)

\bibitem{Yamamoto1998}
R.~Yamamoto, A.~Onuki, Dynamics of highly supercooled liquids: heterogeneity,
  rheology, and diffusion.
\newblock Phys. Rev. E \textbf{58}(3), 3515 (1998)

\bibitem{Chepizhko2013}
O.~Chepizhko, F.~Peruani, Diffusion, subdiffusion, and trapping of active
  particles in heterogeneous media.
\newblock Phys. Rev. Lett. \textbf{111}(16), 160604 (2013)

\bibitem{Yamamoto1998Heterogeneous}
R.~Yamamoto, A.~Onuki, Heterogeneous diffusion in highly supercooled liquids.
\newblock Phys. Rev. Lett. \textbf{81}(22), 4915 (1998)

\bibitem{Lanoiselee2018}
Y.~Lanoisel{\'e}e, N.~Moutal, D.S. Grebenkov, Diffusion-limited reactions in
  dynamic heterogeneous media.
\newblock Nat. Commun. \textbf{9}(1), 4398 (2018)

\bibitem{Hurtado2007}
P.I. Hurtado, L.~Berthier, W.~Kob, Heterogeneous diffusion in a reversible gel.
\newblock Phys. Rev. Lett. \textbf{98}(13), 135503 (2007)

\bibitem{Cicerone1997}
M.T. Cicerone, P.A. Wagner, M.D. Ediger, Translational diffusion on
  heterogeneous lattices: a model for dynamics in glass forming materials.
\newblock J. Phys. Chem. B \textbf{101}(43), 8727--8734 (1997)

\bibitem{Cherstvy2013}
A.G. Cherstvy, A.V. Chechkin, R.~Metzler, Anomalous diffusion and ergodicity
  breaking in heterogeneous diffusion processes.
\newblock New J. Phys. \textbf{15}(8), 083039 (2013)

\bibitem{weigel2011ergodic}
A.V. Weigel, B.~Simon, M.M. Tamkun, D.~Krapf, Ergodic and nonergodic processes
  coexist in the plasma membrane as observed by single-molecule tracking.
\newblock Proc. Natl. Acad. Sci. USA \textbf{108}(16), 6438--6443 (2011)

\bibitem{manzo2015weak}
C.~Manzo, J.A. Torreno-Pina, P.~Massignan, G.J. Lapeyre, M.~Lewenstein, M.F.
  Garcia~Parajo, Weak ergodicity breaking of receptor motion in living cells
  stemming from random diffusivity.
\newblock Phys. Rev. X \textbf{5}(1), 011021 (2015)

\bibitem{Klafter2}
J.~Klafter, G.~Zumofen, L{\'e}vy statistics in a {H}amiltonian system.
\newblock Phys. Rev. E \textbf{49}(6), 4873 (1994)

\bibitem{Metzler2019}
R.~Metzler, Brownian motion and beyond: first-passage, power spectrum,
  non-{G}aussianity, and anomalous diffusion.
\newblock J. Stat. Mech. \textbf{2019}(11), 114003 (2019)

\bibitem{Vilk2022}
O.~Vilk, E.~Aghion, R.~Nathan, S.~Toledo, R.~Metzler, M.~Assaf, Classification
  of anomalous diffusion in animal movement data using power spectral analysis.
\newblock J. Phys. A: Math. Theor. \textbf{55}(33), 334004 (2022)

\bibitem{Krapf2019}
D.~Krapf, N.~Lukat, E.~Marinari, R.~Metzler, G.~Oshanin, C.~Selhuber-Unkel,
  A.~Squarcini, L.~Stadler, M.~Weiss, X.~Xu, Spectral content of a single
  non-{B}rownian trajectory.
\newblock Phys. Rev. X \textbf{9}(1), 011019 (2019)

\bibitem{Burov2011}
S.~Burov, J.H. Jeon, R.~Metzler, E.~Barkai, Single particle tracking in systems
  showing anomalous diffusion: the role of weak ergodicity breaking.
\newblock Phys. Chem. Chem. Phys. \textbf{13}(5), 1800--1812 (2011)

\bibitem{Sabzikar2022}
F.~Sabzikar, J.~Kabala, K.~Burnecki, Tempered fractionally integrated process
  with stable noise as a transient anomalous diffusion model.
\newblock J. Phys. A: Math. Theor. \textbf{55}(17), 174002 (2022)

\bibitem{Liu2017}
K.~Liu, Y.~Chen, X.~Zhang, An evaluation of {ARFIMA} (autoregressive fractional
  integral moving average) programs.
\newblock Axioms \textbf{6}(2), 16 (2017)

\bibitem{Burnecki2015}
K.~Burnecki, E.~Kepten, Y.~Garini, G.~Sikora, A.~Weron, Estimating the
  anomalous diffusion exponent for single particle tracking data with
  measurement errors-an alternative approach.
\newblock Sci. Rep. \textbf{5}(1), 11306 (2015)

\bibitem{Magdziarz2009}
M.~Magdziarz, A.~Weron, K.~Burnecki, J.~Klafter, Fractional {B}rownian motion
  versus the continuous-time random walk: a simple test for subdiffusive
  dynamics.
\newblock Phys. Rev. Lett. \textbf{103}(18), 180602 (2009)

\bibitem{Meyer2022}
P.G. Meyer, E.~Aghion, H.~Kantz, Decomposing the effect of anomalous diffusion
  enables direct calculation of the {H}urst exponent and model classification
  for single random paths.
\newblock J. Phys. A: Math. Theor. \textbf{55}(27), 274001 (2022)

\bibitem{Maraj2020}
K.~Maraj, D.~Szarek, G.~Sikora, A.~Wy{\l}oma{\'n}ska, Empirical anomaly measure
  for finite-variance processes.
\newblock J. Phys. A: Math. Theor. \textbf{54}(2), 024001 (2020)

\bibitem{Magdziarz2020}
M.~Magdziarz, T.~Zorawik, Limit properties of {L}{\'e}vy walks.
\newblock J. Phys. A: Math. Theor. \textbf{53}(50), 504001 (2020)

\bibitem{Wang2020}
W.~Wang, A.G. Cherstvy, A.V. Chechkin, S.~Thapa, F.~Seno, X.~Liu, R.~Metzler,
  Fractional {B}rownian motion with random diffusivity: emerging residual
  nonergodicity below the correlation time.
\newblock J. Phys. A: Math. Theor. \textbf{53}(47), 474001 (2020)

\bibitem{Bhowmik2018}
B.P. Bhowmik, I.~Tah, S.~Karmakar, Non-{G}aussianity of the van {H}ove function
  and dynamic-heterogeneity length scale.
\newblock Phys. Rev. E \textbf{98}(2), 022122 (2018)

\bibitem{Katz1985}
M.J. Katz, E.B. George, Fractals and the analysis of growth paths.
\newblock Bull. Math. Biol. \textbf{47}(2), 273--286 (1985)

\bibitem{Tejedor2010}
V.~Tejedor, O.~B{\'e}nichou, R.~Voituriez, R.~Jungmann, F.~Simmel,
  C.~Selhuber-Unkel, L.B. Oddershede, R.~Metzler, Quantitative analysis of
  single particle trajectories: mean maximal excursion method.
\newblock Biophys. J. \textbf{98}(7), 1364--1372 (2010)

\bibitem{Ernst2014}
D.~Ernst, J.~K{\"o}hler, M.~Weiss, Probing the type of anomalous diffusion with
  single-particle tracking.
\newblock Phys. Chem. Chem. Phys. \textbf{16}(17), 7686--7691 (2014)

\bibitem{Seckler2023}
H.~Seckler, J.~Szwabinski, R.~Metzler, Machine-learning solutions for the
  analysis of single-particle diffusion trajectories.
\newblock J. Phys. Chem. Lett. \textbf{14}(35), 7910--7923 (2023)

\bibitem{He2016}
K.~He, X.~Zhang, S.~Ren, J.~Sun, Deep residual learning for image recognition.
\newblock Proc. of the IEEE Conf. on Computer Vision and Pattern Recognition
  pp. 770--778 (2016)

\bibitem{Szegedy2015}
C.~Szegedy, W.~Liu, Y.~Jia, P.~Sermanet, S.~Reed, D.~Anguelov, D.~Erhan,
  V.~Vanhoucke, A.~Rabinovich, Going deeper with convolutions.
\newblock Proc. of the IEEE Conf. on Computer Vision and Pattern Recognition
  pp. 1--9 (2015)

\bibitem{Kiranyaz2021}
S.~Kiranyaz, O.~Avci, O.~Abdeljaber, T.~Ince, M.~Gabbouj, D.J. Inman, 1{D}
  convolutional neural networks and applications: a survey.
\newblock Mech. Syst. Signal Process. \textbf{151}, 107398 (2021)

\bibitem{Hochreiter1997}
S.~Hochreiter, J.~Schmidhuber, Long short-term memory.
\newblock Neural Comput.  (1997)

\bibitem{Zhai2018}
J.~Zhai, S.~Zhang, J.~Chen, Q.~He, Autoencoder and its various variants.
\newblock Proc. IEEE Int. Conf. Syst. Man Cybern. (SMC) pp. 415--419 (2018)

\bibitem{Chen2016}
T.~Chen, C.~Guestrin, {XGBoost}: a scalable tree boosting system.
\newblock Proc. 22nd ACM SIGKDD Int. Conf. Knowl. Discov. Data Min. pp.
  785--794 (2016)

\bibitem{Oord2016}
A.~van~den Oord, S.~Dieleman, H.~Zen, K.~Simonyan, O.~Vinyals, A.~Graves,
  N.~Kalchbrenner, A.~Senior, K.~Kavukcuoglu, Wave{N}et: a generative model for
  raw audio.
\newblock arXiv:1609.03499  (2016)

\bibitem{Vaswani2017}
A.~Vaswani~{et al.}, Attention is all you need.
\newblock Proc. of the Int. Conf. on Neural Information Processing Systems
  (2017)

\bibitem{Li2021cnn}
Z.~Li, F.~Liu, W.~Yang, S.~Peng, J.~Zhou, A survey of convolutional neural
  networks: analysis, applications, and prospects.
\newblock IEEE Trans. Neural Netw. Learn. Syst. \textbf{33}(12), 6999--7019
  (2021)

\bibitem{Yu2019}
Y.~Yu, X.~Si, C.~Hu, J.~Zhang, A review of recurrent neural networks: {LSTM}
  cells and network architectures.
\newblock Neural Comput. \textbf{31}(7), 1235--1270 (2019)

\bibitem{Scarselli2008}
F.~Scarselli, M.~Gori, A.C. Tsoi, M.~Hagenbuchner, G.~Monfardini, The graph
  neural network model.
\newblock IEEE Trans. Neural Netw. \textbf{20}(1), 61--80 (2008)

\bibitem{Tavenard2020}
R.~Tavenard, J.~Faouzi, G.~Vandewiele, F.~Divo, G.~Androz, C.~Holtz, M.~Payne,
  R.~Yurchak, M.~Ru{\ss}wurm, K.~Kolar~{et al.}, Tslearn, a machine learning
  toolkit for time series data.
\newblock J. Mach. Learn. Res. \textbf{21}(118), 1--6 (2020)

\bibitem{Gruver2024}
N.~Gruver, M.~Finzi, S.~Qiu, A.G. Wilson, Large language models are zero-shot
  time series forecasters.
\newblock Adv. Neural Inf. Process. Syst. \textbf{36} (2024)

\bibitem{Worden2007}
K.~Worden, G.~Manson, The application of machine learning to structural health
  monitoring.
\newblock Philos. Trans. R. Soc. A \textbf{365}(1851), 515--537 (2007)

\bibitem{Ballard2021}
Z.~Ballard, C.~Brown, A.M. Madni, A.~Ozcan, Machine learning and
  computation-enabled intelligent sensor design.
\newblock Nat. Mach. Intell. \textbf{3}(7), 556--565 (2021)

\bibitem{andi_website}
An{D}i: the anomalous diffusion challenge.
\newblock \url{http://andi-challenge.org}

\bibitem{Thapa2018}
S.~Thapa, M.A. Lomholt, J.~Krog, A.G. Cherstvy, R.~Metzler, Bayesian analysis
  of single-particle tracking data using the nested-sampling algorithm:
  maximum-likelihood model selection applied to stochastic-diffusivity data.
\newblock Phys. Chem. Chem. Phys. \textbf{20}(46), 29018--29037 (2018)

\bibitem{Park2021}
S.~Park, S.~Thapa, Y.~Kim, M.A. Lomholt, J.H. Jeon, Bayesian inference of
  {L}{\'e}vy walks via hidden {M}arkov models.
\newblock J. Phys. A: Math. Theor. \textbf{54}(48), 484001 (2021)

\bibitem{Thapa2022}
S.~Thapa, S.~Park, Y.~Kim, J.H. Jeon, R.~Metzler, M.A. Lomholt, Bayesian
  inference of scaled versus fractional {B}rownian motion.
\newblock J. Phys. A: Math. Theor. \textbf{55}(19), 194003 (2022)

\bibitem{Chen2022}
Z.~Chen, L.~Geffroy, J.~Biteen, {NOBIAS}: analyzing anomalous diffusion in
  single-molecule tracks with nonparametric {B}ayesian inference.
\newblock Biophys. J. \textbf{121}(3), 20a (2022)

\bibitem{Krog2017}
J.~Krog, M.A. Lomholt, Bayesian inference with information content model check
  for {L}angevin equations.
\newblock Phys. Rev.E \textbf{96}(6), 062106 (2017)

\bibitem{Krog}
J.~Krog, L.H. Jacobsen, F.W. Lund, D.~W{\"u}stner, M.A. Lomholt, Bayesian model
  selection with fractional {B}rownian motion.
\newblock J. Stat. Mech. \textbf{2018}(9), 093501 (2018)

\bibitem{Manzo2015spt}
C.~Manzo, M.F. Garcia-Parajo, A review of progress in single particle tracking:
  from methods to biophysical insights.
\newblock Rep. Prog. Phys. \textbf{78}(12), 124601 (2015)

\bibitem{Shen2017}
H.~Shen, L.J. Tauzin, R.~Baiyasi, W.~Wang, N.~Moringo, B.~Shuang, C.F. Landes,
  Single particle tracking: from theory to biophysical applications.
\newblock Chem. Rev. \textbf{117}(11), 7331--7376 (2017)

\bibitem{Qian1991}
H.~Qian, M.P. Sheetz, E.L. Elson, Single particle tracking. {A}nalysis of
  diffusion and flow in two-dimensional systems.
\newblock Biophys. J. \textbf{60}(4), 910--921 (1991)

\bibitem{Saxton2008}
M.J. Saxton, Single-particle tracking: connecting the dots.
\newblock Nat. Meth. \textbf{5}(8), 671--672 (2008)

\bibitem{Torreno-Pina2016}
J.A. Torreno-Pina, C.~Manzo, M.F. Garcia-Parajo, Uncovering homo-and
  hetero-interactions on the cell membrane using single particle tracking
  approaches.
\newblock J. Phys. D: Appl. Phys. \textbf{49}(10), 104002 (2016)

\bibitem{Elf2019}
J.~Elf, I.~Barkefors, Single-molecule kinetics in living cells.
\newblock Annu. Rev. Biochem. \textbf{88}(1), 635--659 (2019)

\bibitem{Cherstvy2019}
A.G. Cherstvy, S.~Thapa, C.E. Wagner, R.~Metzler, Non-{G}aussian, non-ergodic,
  and non-{F}ickian diffusion of tracers in mucin hydrogels.
\newblock Soft Matter \textbf{15}(12), 2526--2551 (2019)

\bibitem{Horton2010}
M.R. Horton, F.~H{\"o}fling, J.O. R{\"a}dler, T.~Franosch, Development of
  anomalous diffusion among crowding proteins.
\newblock Soft Matter \textbf{6}(12), 2648--2656 (2010)

\bibitem{Jeon2011}
J.H. Jeon, V.~Tejedor, S.~Burov, E.~Barkai, C.~Selhuber-Unkel,
  K.~Berg-S{\o}rensen, L.~Oddershede, R.~Metzler, In vivo anomalous diffusion
  and weak ergodicity breaking of lipid granules.
\newblock Phys. Rev. Lett. \textbf{106}(4), 048103 (2011)

\bibitem{Leijnse2012}
N.~Leijnse, J.H. Jeon, S.~Loft, R.~Metzler, L.B. Oddershede, Diffusion inside
  living human cells.
\newblock Biophys. J. \textbf{102}(3), 377a (2012)

\bibitem{Codling2008}
E.A. Codling, M.J. Plank, S.~Benhamou, Random walk models in biology.
\newblock J. R. Soc. Interface \textbf{5}(25), 813--834 (2008)

\bibitem{Gurtovenko2019}
A.A. Gurtovenko, M.~Javanainen, F.~Lolicato, I.~Vattulainen, The devil is in
  the details: what do we really track in single-particle tracking experiments
  of diffusion in biological membranes?
\newblock J. Phys. Chem. Lett. \textbf{10}(5), 1005--1011 (2019)

\bibitem{Smal2009}
I.~Smal, M.~Loog, W.~Niessen, E.~Meijering, Quantitative comparison of spot
  detection methods in fluorescence microscopy.
\newblock IEEE Trans. Med. Imaging \textbf{29}(2), 282--301 (2009)

\bibitem{Roberts2020}
T.D. Roberts, R.~Yuan, L.~Xiang, M.~Delor, R.~Pokhrel, K.~Yang, E.~Aqad,
  T.~Marangoni, P.~Trefonas, K.~Xu~{et al.}, Direct correlation of
  single-particle motion to amorphous microstructural components of
  semicrystalline poly (ethylene oxide) electrolytic films.
\newblock J. Phys. Chem. Lett. \textbf{11}(12), 4849--4858 (2020)

\bibitem{Ye2023}
Z.~Ye, C.~Hu, J.~Wang, H.~Liu, L.~Li, J.~Yuan, J.W. Ha, Z.~Li, L.~Xiao, Burst
  of hopping trafficking correlated reversible dynamic interactions between
  lipid droplets and mitochondria under starvation.
\newblock Exploration \textbf{3}(5), 20230002 (2023)

\bibitem{Erimban2023}
S.~Erimban, S.~Daschakraborty, Fickian yet non-{G}aussian nanoscopic lipid
  diffusion in the raft-mimetic membrane.
\newblock J. Phys. Chem. B \textbf{127}(22), 4939--4951 (2023)

\bibitem{Javanainen2017}
M.~Javanainen, H.~Martinez-Seara, R.~Metzler, I.~Vattulainen, Diffusion of
  integral membrane proteins in protein-rich membranes.
\newblock J. Phys. Chem. Lett. \textbf{8}(17), 4308--4313 (2017)

\bibitem{Winkler2018}
P.M. Winkler, R.~Regmi, V.~Flauraud, J.~Brugger, H.~Rigneault, J.~Wenger, M.F.
  Garc{\'\i}a-Parajo, Optical antenna-based fluorescence correlation
  spectroscopy to probe the nanoscale dynamics of biological membranes.
\newblock J. Phys. Chem. Lett. \textbf{9}(1), 110--119 (2018)

\bibitem{Simon2024}
F.~Simon, L.E. Weiss, S.~van Teeffelen, A guide to single-particle tracking.
\newblock Nat. Rev. Methods Primers \textbf{4}(1), 66 (2024)

\bibitem{Munoz-Gil2}
G.~Mu{\~n}oz-Gil, M.A. Garcia-March, C.~Manzo, J.D. Mart{\'\i}n-Guerrero,
  M.~Lewenstein, Single trajectory characterization via machine learning.
\newblock New J. Phys. \textbf{22}(1), 013010 (2020)

\bibitem{Kowalek2022}
P.~Kowalek, H.~Loch-Olszewska, {\L}.~{\L}aszczuk, J.~Opa{\l}a,
  J.~Szwabi{\'n}ski, Boosting the performance of anomalous diffusion
  classifiers with the proper choice of features.
\newblock J. Phys. A: Math. Theor. \textbf{55}(24), 244005 (2022)

\bibitem{Manzo2021}
C.~Manzo, Extreme learning machine for the characterization of anomalous
  diffusion from single trajectories ({AnDi-ELM}).
\newblock J. Phys. A: Math. Theor. \textbf{54}(33), 334002 (2021)

\bibitem{Loch2020}
H.~Loch-Olszewska, J.~Szwabi{\'n}ski, Impact of feature choice on machine
  learning classification of fractional anomalous diffusion.
\newblock Entropy \textbf{22}(12), 1436 (2020)

\bibitem{Gajowczyk2021}
M.~Gajowczyk, J.~Szwabi{\'n}ski, Detection of anomalous diffusion with deep
  residual networks.
\newblock Entropy \textbf{23}(6), 649 (2021)

\bibitem{Granik2019}
N.~Granik, L.E. Weiss, E.~Nehme, M.~Levin, M.~Chein, E.~Perlson, Y.~Roichman,
  Y.~Shechtman, Single-particle diffusion characterization by deep learning.
\newblock Biophys. J. \textbf{117}(2), 185--192 (2019)

\bibitem{AL-hada2022}
E.A. AL-hada, X.~Tang, W.~Deng, Classification of stochastic processes by
  convolutional neural networks.
\newblock J. Phys. A: Math. Theor. \textbf{55}(27), 274006 (2022)

\bibitem{Conejero2023}
J.A. Conejero, {\`O}.~{Garibo-i-Orts}, C.~Lizama, Inferring the fractional
  nature of {W}u {B}aleanu trajectories.
\newblock Nonlinear Dyn. \textbf{111}(13), 12421--12431 (2023)

\bibitem{Li2021}
D.~Li, Q.~Yao, Z.~Huang, Wave{N}et-based deep neural networks for the
  characterization of anomalous diffusion ({WADN}et).
\newblock J. Phys. A: Math. Theor. \textbf{54}(40), 404003 (2021)

\bibitem{DeepSPT2020}
T.~Song.
\newblock Deep{SPT}. (2020).
\newblock \url{https://github.com/AnDiChallenge/AnDi2020_TeamD_DeepSPT}

\bibitem{NOA2020}
N.~Firbas.
\newblock {NOA}. (2020).
\newblock \url{https://github.com/AnDiChallenge/AnDi2020_TeamH_NOA}

\bibitem{Firbas2023}
N.~Firbas, {\`O}.~{Garibo-i-Orts}, M.{\'A}. Garcia-March, J.A. Conejero,
  Characterization of anomalous diffusion through convolutional transformers.
\newblock J. Phys. A: Math. Theor. \textbf{56}(1), 014001 (2023)

\bibitem{Feng2024}
X.~Feng, H.~Sha, Y.~Zhang, Y.~Su, S.~Liu, Y.~Jiang, S.~Hou, S.~Han, X.~Ji,
  Reliable deep learning in anomalous diffusion against out-of-distribution
  dynamics.
\newblock Nat. Comput. Sci pp. 1--12 (2024)

\bibitem{Bo2019}
S.~Bo, F.~Schmidt, R.~Eichhorn, G.~Volpe, Measurement of anomalous diffusion
  using recurrent neural networks.
\newblock Phys. Rev. E \textbf{100}(1), 010102 (2019)

\bibitem{Argun2021}
A.~Argun, G.~Volpe, S.~Bo, Classification, inference and segmentation of
  anomalous diffusion with recurrent neural networks.
\newblock J. Phys. A: Math. Theor. \textbf{54}(29), 294003 (2021)

\bibitem{Garibo-i-Orts2021}
{\`O}.~{Garibo-i-Orts}, A.~Baeza-Bosca, M.A. Garcia-March, J.A. Conejero,
  Efficient recurrent neural network methods for anomalously diffusing single
  particle short and noisy trajectories.
\newblock J. Phys. A: Math. Theor. \textbf{54}(50), 504002 (2021)

\bibitem{Kabbech2024}
H.~Kabbech, I.~Smal, {TrackSegNet}: a tool for trajectory segmentation into
  diffusive states using supervised deep learning.
\newblock J. Open Source Softw. \textbf{9}(98), 6157 (2024)

\bibitem{Verdier2021}
H.~Verdier, M.~Duval, F.~Laurent, A.~Cass{\'e}, C.L. Vestergaard, J.B. Masson,
  Learning physical properties of anomalous random walks using graph neural
  networks.
\newblock J. Phys. A: Math. Theor. \textbf{54}(23), 234001 (2021)

\bibitem{Verdier2022}
H.~Verdier, F.~Laurent, A.~Cass{\'e}, C.L. Vestergaard, J.B. Masson,
  Variational inference of fractional {B}rownian motion with linear
  computational complexity.
\newblock Phys. Rev. E \textbf{106}(5), 055311 (2022)

\bibitem{Pineda2023}
J.~Pineda, B.~Midtvedt, H.~Bachimanchi, S.~No{\'e}, D.~Midtvedt, G.~Volpe,
  C.~Manzo, Geometric deep learning reveals the spatiotemporal features of
  microscopic motion.
\newblock Nat. Mach. Intell. \textbf{5}(1), 71--82 (2023)

\bibitem{Seckler2022}
H.~Seckler, R.~Metzler, Bayesian deep learning for error estimation in the
  analysis of anomalous diffusion.
\newblock Nat. Commun. \textbf{13}(1), 6717 (2022)

\bibitem{QuSoftware}
X.~Qu, H.~Zhao, W.~Cai, G.~Wang, Z.~Huang, {GenML}: a {Python} library to
  generate the {Mittag-Leffler} correlated noise.
\newblock arXiv:2403.04273  (2024)

\bibitem{golding2006physical}
I.~Golding, E.C. Cox, Physical nature of bacterial cytoplasm.
\newblock Phys. Rev. Lett. \textbf{96}(9), 098102 (2006)

\bibitem{stadler2017non}
L.~Stadler, M.~Weiss, Non-equilibrium forces drive the anomalous diffusion of
  telomeres in the nucleus of mammalian cells.
\newblock New J. Phys. \textbf{19}(11), 113048 (2017)

\bibitem{kindermann2017nonergodic}
F.~Kindermann, A.~Dechant, M.~Hohmann, T.~Lausch, D.~Mayer, F.~Schmidt,
  E.~Lutz, A.~Widera, Nonergodic diffusion of single atoms in a periodic
  potential.
\newblock Nat. Phys. \textbf{13}(2), 137--141 (2017)

\bibitem{Meyes2019}
R.~Meyes, M.~Lu, C.W. de~Puiseau, T.~Meisen, Ablation studies in artificial
  neural networks.
\newblock arXiv:1901.08644  (2019)

\bibitem{Qu2024}
X.~Qu, Y.~Hu, W.~Cai, Y.~Xu, H.~Ke, G.~Zhu, Z.~Huang, Semantic segmentation of
  anomalous diffusion using deep convolutional networks.
\newblock Phys. Rev. Res. \textbf{6}(1), 013054 (2024)

\bibitem{Vega2018}
A.R. Vega, S.A. Freeman, S.~Grinstein, K.~Jaqaman, Multistep track segmentation
  and motion classification for transient mobility analysis.
\newblock Biophys. J. \textbf{114}(5), 1018--1025 (2018)

\bibitem{Dosset2016}
P.~Dosset, P.~Rassam, L.~Fernandez, C.~Espenel, E.~Rubinstein, E.~Margeat, P.E.
  Milhiet, Automatic detection of diffusion modes within biological membranes
  using back-propagation neural network.
\newblock BMC Bioinform. \textbf{17}, 1--12 (2016)

\bibitem{Zhang2023}
Y.~Zhang, F.~Ge, X.~Lin, J.~Xue, Y.~Song, H.~Xie, Y.~He, Extract latent
  features of single-particle trajectories with historical experience learning.
\newblock Biophys. J. \textbf{122}(22), 4451--4466 (2023)

\bibitem{Yu2014}
Y.~Yu, Y.~Zhu, S.~Li, D.~Wan, Time series outlier detection based on sliding
  window prediction.
\newblock Math. Probl. Eng. \textbf{2014}(1), 879736 (2014)

\bibitem{Helmuth2007}
J.A. Helmuth, C.J. Burckhardt, P.~Koumoutsakos, U.F. Greber, I.F. Sbalzarini, A
  novel supervised trajectory segmentation algorithm identifies distinct types
  of human adenovirus motion in host cells.
\newblock J. Struct. Biol. \textbf{159}(3), 347--358 (2007)

\bibitem{Weron2017}
A.~Weron, K.~Burnecki, E.J. Akin, L.~Sol{\'e}, M.~Balcerek, M.M. Tamkun,
  D.~Krapf, Ergodicity breaking on the neuronal surface emerges from random
  switching between diffusive states.
\newblock Sci. Rep. \textbf{7}(1), 5404 (2017)

\bibitem{Sikora2017}
G.~Sikora, A.~Wy{\l}oma{\'n}ska, J.~Gajda, L.~Sol{\'e}, E.J. Akin, M.M. Tamkun,
  D.~Krapf, Elucidating distinct ion channel populations on the surface of
  hippocampal neurons via single-particle tracking recurrence analysis.
\newblock Phys. Rev. E \textbf{96}(6), 062404 (2017)

\bibitem{Requena2023}
B.~Requena, S.~Mas{\'o}-Orriols, J.~Bertran, M.~Lewenstein, C.~Manzo,
  G.~Mu{\~n}oz-Gil, Inferring pointwise diffusion properties of single
  trajectories with deep learning.
\newblock Biophys. J. \textbf{122}(22), 4360--4369 (2023)

\bibitem{Munoz2023}
G.~Mu{\~n}oz-Gil, H.~Bachimanchi, J.~Pineda, B.~Midtvedt, M.~Lewenstein,
  R.~Metzler, D.~Krapf, G.~Volpe, C.~Manzo, Quantitative evaluation of methods
  to analyze motion changes in single-particle experiments.
\newblock arXiv:2311.18100  (2023)

\bibitem{Matsuoka2009}
S.~Matsuoka, T.~Shibata, M.~Ueda, Statistical analysis of lateral diffusion and
  multistate kinetics in single-molecule imaging.
\newblock Biophys. J. \textbf{97}(4), 1115--1124 (2009)

\bibitem{Wagner2017}
T.~Wagner, A.~Kroll, C.R. Haramagatti, H.G. Lipinski, M.~Wiemann,
  Classification and segmentation of nanoparticle diffusion trajectories in
  cellular micro environments.
\newblock PloS One \textbf{12}(1), e0170165 (2017)

\bibitem{Matsuda2018}
Y.~Matsuda, I.~Hanasaki, R.~Iwao, H.~Yamaguchi, T.~Niimi, Estimation of
  diffusive states from single-particle trajectory in heterogeneous medium
  using machine-learning methods.
\newblock Phys. Chem. Chem. Phys. \textbf{20}(37), 24099--24108 (2018)

\bibitem{Gentili2021}
A.~Gentili, G.~Volpe, Characterization of anomalous diffusion classical
  statistics powered by deep learning ({CONDOR}).
\newblock J. Phys. A: Math. Theor. \textbf{54}(31), 314003 (2021)

\bibitem{Arts2019}
M.~Arts, I.~Smal, M.W. Paul, C.~Wyman, E.~Meijering, Particle mobility analysis
  using deep learning and the moment scaling spectrum.
\newblock Sci. Rep. \textbf{9}(1), 17160 (2019)

\bibitem{Martinez2023}
Q.~Martinez, C.~Chen, J.~Xia, H.~Bahai, Sequence-to-sequence change-point
  detection in single-particle trajectories via recurrent neural network for
  measuring self-diffusion.
\newblock Transp. Porous Med. \textbf{147}(3), 679--701 (2023)

\bibitem{seckler2024change}
H.~Seckler, R.~Metzler, Change-point detection in anomalous-diffusion
  trajectories utilising machine-learning-based uncertainty estimates.
\newblock J. Phys. Photonics \textbf{6}(4), 045025 (2024)

\bibitem{Mo2022}
Y.~Mo, Y.~Wu, X.~Yang, F.~Liu, Y.~Liao, Review the state-of-the-art
  technologies of semantic segmentation based on deep learning.
\newblock Neurocomputing \textbf{493}, 626--646 (2022)

\bibitem{Deng2009}
J.~Deng, W.~Dong, R.~Socher, L.J. Li, K.~Li, L.~Fei-Fei, Imagenet: a
  large-scale hierarchical image database.
\newblock Proc. IEEE Conf. Computer Vision and Pattern Recognition pp. 248--255
  (2009)

\bibitem{Lin2014}
T.Y. Lin, M.~Maire, S.~Belongie, J.~Hays, P.~Perona, D.~Ramanan, P.~Doll{\'a}r,
  C.L. Zitnick, Microsoft coco: common objects in context.
\newblock Proc. Eur. Conf. Comput. Vis. (ECCV) pp. 740--755 (2014)

\bibitem{kaggle}
Kaggle.
\newblock \url{https://www.kaggle.com}

\bibitem{munoz2020anomalous}
G.~Mu{\~n}oz-Gil, G.~Volpe, M.A. Garc{\'\i}a-March, R.~Metzler, M.~Lewenstein,
  C.~Manzo, \emph{The anomalous diffusion challenge: single trajectory
  characterisation as a competition}, in \emph{Emerging Topics in Artificial
  Intelligence 2020}, vol. 11469 (SPIE, 2020), pp. 42--51

\bibitem{codalab}
A.~Pavao, I.~Guyon, A.C. Letournel, D.T. Tran, X.~Baro, H.J. Escalante,
  S.~Escalera, T.~Thomas, Z.~Xu, Codalab competitions: An open source platform
  to organize scientific challenges.
\newblock J. Mach. Learn. Res. \textbf{24}(198), 1--6 (2023).
\newblock \url{http://jmlr.org/papers/v24/21-1436.html}

\bibitem{AnDi-unicorns}
B.~Requena.
\newblock Anomalous {U}nicorns. (2020).
\newblock
  \url{https://github.com/AnDiChallenge/AnDi2020_TeamA_AnomalousUnicorns}

\bibitem{ErasmusMC}
H.~Kabbech.
\newblock Erasmus {MC}. (2020).
\newblock \url{https://github.com/AnDiChallenge/AnDi2020_TeamF_ErasmusMC}

\bibitem{FCI}
T.~Bland.
\newblock {FCI}. (2020).
\newblock \url{https://github.com/AnDiChallenge/AnDi2020_TeamJ_FCI}

\bibitem{Aghion}
E.~Aghion, P.G. Meyer, V.~Adlakha, H.~Kantz, K.E. Bassler, Moses, {N}oah and
  {J}oseph effects in {L}{\'e}vy walks.
\newblock New J. Phys. \textbf{23}(2), 023002 (2021)

\bibitem{Wust-ML-A}
J.~Szwabi{\'n}ski.
\newblock {Wust ML A}. (2020).
\newblock \url{https://github.com/AnDiChallenge/AnDi2020_TeamN_WustMLA}

\bibitem{Wust-ML-B}
H.~Loch-Olszewska, P.~Kowalek.
\newblock {Wust ML B}. (2020).
\newblock \url{https://github.com/AnDiChallenge/AnDi2020_TeamO_WustMLB1}

\bibitem{Gorka2020}
G.~Mu\~{n}oz Gil, C.~Manzo, G.~Volpe, M.A. Garcia-March, R.~Metzler,
  M.~Lewenstein.
\newblock The anomalous diffusion challenge dataset (2020).
\newblock \url{https://zenodo.org/record/3707702}

\bibitem{Massignan}
P.~Massignan, C.~Manzo, J.A. Torreno-Pina, M.F. Garc{\'\i}a-Parajo,
  M.~Lewenstein, G.J. Lapeyre~Jr, Nonergodic subdiffusion from {B}rownian
  motion in an inhomogeneous medium.
\newblock Phys. Rev. Lett. \textbf{112}(15), 150603 (2014)

\bibitem{Lim}
S.C. Lim, S.V. Muniandy, Self-similar {G}aussian processes for modeling
  anomalous diffusion.
\newblock Phys. Rev. E \textbf{66}(2), 021114 (2002)

\bibitem{Pinholt2021}
H.D. Pinholt, S.S.R. Bohr, J.F. Iversen, W.~Boomsma, N.S. Hatzakis,
  Single-particle diffusional fingerprinting: a machine-learning framework for
  quantitative analysis of heterogeneous diffusion.
\newblock Proc. Natl. Acad. Sci. USA \textbf{118}(31), e2104624118 (2021)

\bibitem{Mangalam2023}
M.~Mangalam, R.~Metzler, D.G. Kelty-Stephen, Ergodic characterization of
  nonergodic anomalous diffusion processes.
\newblock Phys. Rev. Res. \textbf{5}(2), 023144 (2023)

\bibitem{Chhabra1989}
A.~Chhabra, R.V. Jensen, Direct determination of the f ($\alpha$) singularity
  spectrum.
\newblock Phys. Rev. Lett. \textbf{62}(12), 1327 (1989)

\bibitem{Seckler2024}
H.~Seckler, R.~Metzler, D.G. Kelty-Stephen, M.~Mangalam, Multifractal spectral
  features enhance classification of anomalous diffusion.
\newblock Phys. Rev. E \textbf{109}(4), 044133 (2024)

\bibitem{Bengio2013}
Y.~Bengio, A.~Courville, P.~Vincent, Representation learning: a review and new
  perspectives.
\newblock IEEE Trans. Pattern Anal. Mach. Intell. \textbf{35}(8), 1798--1828
  (2013)

\bibitem{Zhong2016}
G.~Zhong, L.N. Wang, X.~Ling, J.~Dong, An overview on data representation
  learning: from traditional feature learning to recent deep learning.
\newblock J. Finance Data Sci. \textbf{2}(4), 265--278 (2016)

\bibitem{VanderMaaten2008}
L.~van~der Maaten, G.~Hinton, Visualizing data using {t-SNE}.
\newblock J. Mach. Learn. Res. \textbf{9}(11) (2008)

\bibitem{McInnes2018}
L.~McInnes, J.~Healy, J.~Melville, {UMAP}: uniform manifold approximation and
  projection for dimension reduction.
\newblock arXiv:1802.03426  (2018)

\bibitem{Crispin2009}
C.W. Gardiner, Handbook of stochastic methods for physics, chemistry, and the
  natural sciences springer.
\newblock Berlin (4th Ed)  (2009)

\bibitem{Fernandez2024}
G.~Fern{\'a}ndez-Fern{\'a}ndez, C.~Manzo, M.~Lewenstein, A.~Dauphin,
  G.~Mu{\~n}oz-Gil, Learning minimal representations of stochastic processes
  with variational autoencoders.
\newblock Phys. Rev. E \textbf{110}(1), L012102 (2024)

\bibitem{MunozUnsupervised}
G.~Mu{\~n}oz-Gil, G.G. i~Corominas, M.~Lewenstein, Unsupervised learning of
  anomalous diffusion data: an anomaly detection approach.
\newblock J. Phys. A: Math. Theor. \textbf{54}(50), 504001 (2021)

\bibitem{andi_dateset}
G.~Mu{\~n}oz-Gil, B.~Requena, G.~Fern{\'a}ndez-Fern{\'a}ndez, H.~Bachimanchi,
  J.~Pineda, C.~Manzo.
\newblock Andichallenge/andi\_datasets: Andi challenge 2 (2023).
\newblock \doi{10.5281/zenodo.10259556}

\end{thebibliography}

\end{document}